\documentclass{article} % For LaTeX2e
\usepackage{roboatom,times}
\usepackage[T1]{fontenc}

\usepackage{amsmath,amsfonts,bm}

\def\eqref#1{equation~\ref{#1}}
\def\1{\bm{1}}

\DeclareMathAlphabet{\mathsfit}{\encodingdefault}{\sfdefault}{m}{sl}
\SetMathAlphabet{\mathsfit}{bold}{\encodingdefault}{\sfdefault}{bx}{n}

\usepackage{url}

\usepackage{amsmath,amssymb}
\usepackage{graphicx}
\usepackage{booktabs}
\usepackage{multirow}
\usepackage{listings}
\usepackage{textcomp}
\usepackage{float}
\usepackage{placeins}
\usepackage{hyperref}

\ifdefined\DeclareUnicodeCharacter
\DeclareUnicodeCharacter{2020}{\dag}
\DeclareUnicodeCharacter{2248}{\ensuremath{\approx}}
\DeclareUnicodeCharacter{2192}{\ensuremath{\rightarrow}}
\DeclareUnicodeCharacter{2265}{\ensuremath{\geq}}
\DeclareUnicodeCharacter{2212}{\ensuremath{-}}
\DeclareUnicodeCharacter{00D7}{\ensuremath{\times}}
\DeclareUnicodeCharacter{03C0}{\ensuremath{\pi}}
\DeclareUnicodeCharacter{2460}{\ding{172}}
\DeclareUnicodeCharacter{2461}{\ding{173}}
\DeclareUnicodeCharacter{2462}{\ding{174}}
\DeclareUnicodeCharacter{2463}{\ding{175}}
\fi
\usepackage{pifont}

\definecolor{resourcelinkcolor}{rgb}{0.20,0.29,0.40}
\newcommand{\resourcelink}[3]{%
  \begingroup
  \hypersetup{hidelinks}%
  \href{#1}{\mbox{%
    \raisebox{-2pt}{\includegraphics[height=11pt]{assets/icons/#2.pdf}}%
    \hspace{0.25em}\textcolor{resourcelinkcolor}{#3}}}%
  \endgroup
}

\title{RoboAtom: Large-scale Atomic-Action \\ Supervision for Robot Learning}

\author{%
  \parbox[b]{\dimexpr\textwidth-2\tabcolsep\relax}{%
    \centering\normalfont\bfseries
    Xueyuan~Wei\textsuperscript{1,*},
    Xuanxu~Wang\textsuperscript{5,1,*},
    Xudong~Li\textsuperscript{1,3}, Yuping~He\textsuperscript{1},
    Wanshui~Gan\textsuperscript{1}, Jiaqi~Peng\textsuperscript{4,1},
    Wenqi~Guo\textsuperscript{4,1}, Jianjun~Zhou\textsuperscript{2},
    Junting~Dong\textsuperscript{1}, Chunhua~Shen\textsuperscript{1,2},
    Haoyu~Guo\textsuperscript{1},
    Mulin~Yu\textsuperscript{1,\ensuremath{\dagger}},
    Tao~Lu\textsuperscript{1,\ensuremath{\dagger}}\endgraf
    \medskip
    {\normalfont
      \textsuperscript{1}Shanghai Artificial Intelligence Laboratory\quad
      \textsuperscript{2}Zhejiang University\endgraf
      \textsuperscript{3}Fudan University\quad
      \textsuperscript{4}Tsinghua University\quad
      \textsuperscript{5}Shanghai Jiao Tong University\endgraf}
    \smallskip
    {\normalfont\small
      \textsuperscript{*}Equal contribution\quad
      \textsuperscript{\ensuremath{\dagger}}Corresponding authors\endgraf}
    \smallskip
    {\normalfont\small
      \resourcelink{https://internrobotics.github.io/RoboAtom/}{project}{Project}\qquad
      \resourcelink{https://github.com/InternRobotics/RoboAtom}{github}{Code}\qquad
      \resourcelink{https://huggingface.co/datasets/InternRobotics/RoboAtom}{data}{Data}\qquad
      \resourcelink{https://huggingface.co/InternRobotics/RoboAtom}{models}{Models}\endgraf}
  }%
}

\hypersetup{%
  pdfauthor={Xueyuan Wei, Xuanxu Wang, Xudong Li, Yuping He, Wanshui Gan,
    Jiaqi Peng, Wenqi Guo, Jianjun Zhou, Junting Dong, Chunhua Shen,
    Haoyu Guo, Mulin Yu, Tao Lu},
  pdftitle={RoboAtom: Large-scale Atomic-Action Supervision for Robot Learning}
}

\iclrfinalcopy % Show authors and disable review line numbers for the preprint.
\begin{document}

\maketitle
\lhead{} % Clear the conference publication header set by \maketitle.

% Limit main-paper pages to two floats, at most one per edge.
\edef\restorefloatlimits{%
  \noexpand\setcounter{topnumber}{\arabic{topnumber}}%
  \noexpand\setcounter{bottomnumber}{\arabic{bottomnumber}}%
  \noexpand\setcounter{totalnumber}{\arabic{totalnumber}}%
}
\setcounter{topnumber}{1}
\setcounter{bottomnumber}{1}
\setcounter{totalnumber}{2}

\begin{figure}[h]
\vspace{-1.0cm}
    \centering
    \includegraphics[width=\linewidth]{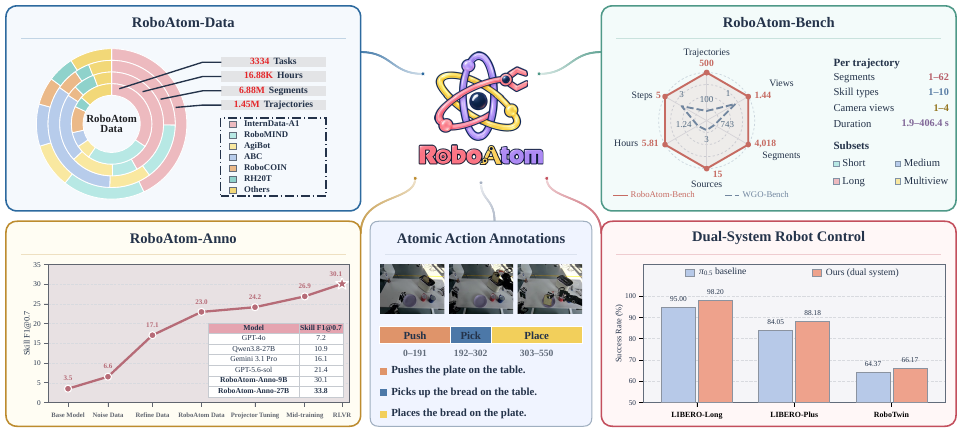}
    \vspace{-0.4cm}
    \caption{Overview of RoboAtom: data and benchmark (top), atomic-action annotations (center), annotation performance (bottom left), and robot-control success rates (bottom right).}
    \label{fig:overview}
\end{figure}

\begin{abstract}
Fine-grained action annotations can support skill reuse and compositional generalization in robot policies. However, robot datasets typically provide task-level instructions without subtask annotations. Although some include subtask labels, inconsistent formats and conventions hinder cross-source joint training. We propose RoboAtom-Data, which adds unified, time-aligned atomic-action annotations to existing data from multiple sources, covering 1.45 million trajectories, 6.88 million skill-labeled segments, and 16,882.1 hours. Our pipeline combines a unified protocol, large-scale manual annotation, propagation across repeated executions with the same ordered skill sequence, and multi-stage quality control. We evaluate the dataset through automatic atomic-action annotation and robot control. We construct RoboAtom-Bench to standardize evaluation of skill recognition and temporal boundary localization. Trained on the annotated data, RoboAtom-Anno achieves better automatic annotation performance than all evaluated open- and closed-source models on both RoboAtom-Bench and the external WGO-Bench. For robot control, atomic-action supervision paired with hierarchical execution improves task success over task-level baselines on LIBERO-Long, LIBERO-Plus, and RoboTwin.  We will release the data, benchmark, model, and construction pipeline to support further research on skill-level robot learning.
\end{abstract}

\section{Introduction}
\label{sec:intro}

Time-aligned atomic-action annotations expose intermediate trajectory structure and can support skill reuse and compositional generalization. For example, ``put the cup into the drawer'' comprises picking up the cup, opening the drawer, and placing the cup inside. Unlike a task-level instruction that specifies the final goal, these annotations attach a skill label, an action description, and temporal boundaries to each action, providing a shared interface for automatic annotation and skill-conditioned execution.

Most robot datasets provide task-level instructions without intermediate-action annotations. Open X-Embodiment~\citep{openx}, DROID~\citep{droid}, RoboMIND~\citep{robomind}, and InternData-A1~\citep{interndata} have expanded coverage across embodiments, tasks, and scenes. Among datasets that do provide subtask annotations, however, annotation formats, vocabularies, granularity, and temporal coverage vary. For example, some RH20T~\citep{rh20t} annotations combine multiple pick-and-place cycles involving different colored blocks into a single segment. These inconsistencies hinder joint training and motivate shared skill definitions and segmentation rules.

Building such annotations at scale is difficult. Trajectory-by-trajectory manual refinement costs about 100 USD per video hour, and general video-language models still struggle to localize brief physical events such as grasps and releases: on RoboAtom-Bench, the best-performing evaluated general video-language model, GPT-5.6-sol, achieves a Skill F1@0.7 of only 21.4. Propagating annotations across repeated executions of the same task is far cheaper, but must tolerate variations in execution speed, pauses, and retries, and must detect trajectories whose action structure differs from the reference.

To address these challenges, we introduce \textbf{RoboAtom}, a unified resource and construction pipeline for time-aligned atomic-action annotations. \textbf{RoboAtom-Data} augments trajectories from 16 sources with standardized skill categories, action descriptions, and temporal boundaries, covering 1.45 million trajectories, 6.88 million segments, and 16,882.1 hours (Table~\ref{tab:datasets}; Figure~\ref{fig:overview}). Its pipeline combines manual seed annotation, \textbf{PhysExtend}, a same-task propagation method that localizes boundaries with skill-specific physical cues, and multi-stage quality assessment and correction.
We validate this supervision through automatic annotation and robot control. \textbf{RoboAtom-Bench} provides skill-aware temporal evaluation for atomic-action annotation. \textbf{RoboAtom-Anno}, trained on the annotated data, achieves better automatic annotation performance than all evaluated open- and closed-source models on the task-disjoint RoboAtom-Bench and the external WGO-Bench~\citep{wgo}. Atomic-action supervision paired with hierarchical execution also improves task success over task-level baselines on LIBERO-Long~\citep{libero}, LIBERO-Plus~\citep{liberoplus}, and RoboTwin~\citep{robotwin}.

The main contributions of this paper are:
{\setlength{\leftmargini}{1.2em}
\begin{itemize}
    \item \textbf{RoboAtom-Data.} We introduce 1.45 million robot trajectories with unified, time-aligned atomic-action annotations, together with a scalable construction pipeline built around PhysExtend, a same-task propagation method guided by skill-specific physical cues.
    \item \textbf{RoboAtom-Bench.} We establish an expert-reviewed benchmark that evaluates skill recognition, temporal boundary localization, and sequence structure across action-sequence lengths and viewing conditions.
    \item \textbf{Downstream validation.} We show that atomic-action supervision yields an annotation model that is both more accurate and far cheaper than general video-language models, and improves robot task success when paired with hierarchical execution.
\end{itemize}
}

\section{Related Work}
\label{sec:related}

\subsection{Robot Datasets and Subtask Annotations}
\label{sec:related-datasets}

Robot datasets support policy generalization through diverse embodiments and environments. Open X-Embodiment~\citep{openx} pools multi-institution experience; BridgeData V2~\citep{bridgedata} and DROID~\citep{droid} emphasize real-world scene and operator diversity; RH20T~\citep{rh20t} adds force and audio; AgiBot World~\citep{agibot} and RoboMIND~\citep{robomind} scale standardized collection; and Galaxea Open-World~\citep{galaxea} covers mobile manipulation. LIBERO~\citep{libero}, RoboCasa~\citep{robocasa}, RoboTwin 2.0~\citep{robotwin}, and InternData-A1~\citep{interndata} provide scalable simulated data through procedural generation and domain randomization.

AgiBot World~\citep{agibot}, Galaxea~\citep{galaxea}, RoboInter-Data~\citep{robointer}, and BEHAVIOR-1K~\citep{behavior1k} also provide subtask annotations (Table~\ref{tab:datasets}), but inconsistent vocabularies, granularity, and temporal coverage prevent direct pooling. For example, some RH20T~\citep{rh20t} annotations combine multiple pick-and-place cycles involving different colored blocks into a single segment. RoboAtom-Data unifies vocabulary and temporal alignment across existing trajectories from 16 sources.

\subsection{Atomic-Action Annotation and Evaluation}
\label{sec:related-automation}

{Atomic-action annotation requires both semantic recognition and temporal localization. Recent work on video temporal grounding, including VTimeLLM~\citep{vtimellm}, TimeChat~\citep{timechat}, TimeLens~\citep{timelens}, and TimeLens2~\citep{timelens2}, studies how multimodal LLMs generate time-localized evidence and improve temporal localization quality. These methods motivate automatic annotation, while our setting additionally requires a robot-specific atomic-action vocabulary and separate evaluation of semantic correctness and temporal boundaries. For robot trajectories, Cortex~\citep{cortex} uses reference segmentations together with state-action and visual features to infer subtask boundaries in other executions. This reuse reduces annotation effort, but variations in execution speed and pauses can complicate boundary alignment.
WGO-Bench~\citep{wgo} evaluates video-to-subtask annotation through temporal segmentation, label semantics, and joint matching. Its results identify temporal localization as a major bottleneck, with brief actions particularly prone to omission. These challenges motivate reliable boundary propagation and evaluation protocols that jointly assess action semantics and temporal accuracy.}

\subsection{Atomic-Action Supervision for Robot Learning}
\label{sec:related-downstream}

{Intermediate representations connect task instructions to continuous actions in hierarchical robot learning. RT-H~\citep{rth}, Embodied Chain-of-Thought~\citep{ecot}, and $\pi_{0.5}$~\citep{pi05} use language motions, intermediate reasoning, or subtask prediction to guide action generation. Hi Robot~\citep{hirobot}, GR00T N1~\citep{groot}, and RoboBrain~\citep{robobrain} separate high-level planning from low-level execution, while Cortex~\citep{cortex} uses canonical skills as its interface. Beyond control, REDS~\citep{reds} and SARM~\citep{sarm} use segmented demonstrations or subtask annotations to learn stage-aware rewards.

These works demonstrate the value of intermediate-action supervision, but typically rely on existing annotations or construct them for specific models and datasets. Cross-source annotation consistency and standalone quality evaluation remain less explored. RoboAtom-Data addresses this upstream need by constructing unified, time-aligned annotations across heterogeneous robot datasets and evaluating their quality and downstream utility.}

\section{RoboAtom-Data}
\label{sec:dataset}

\begin{figure}[t]
    \centering
    \includegraphics[width=0.9\linewidth]{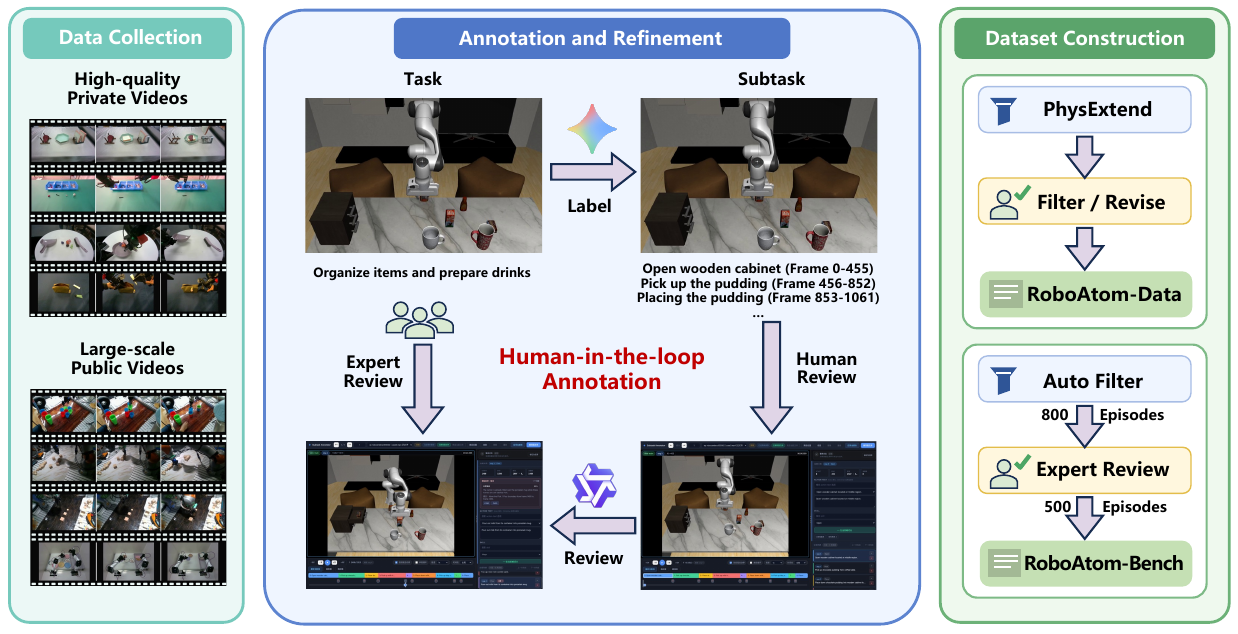}
    \caption{RoboAtom construction pipeline. Model-assisted annotation and human review produce seed annotations; PhysExtend and quality control scale them into RoboAtom-Data. Expert review establishes the evaluation annotations for RoboAtom-Bench (Section~\ref{sec:benchmark}).}
    \label{fig:pipeline}
\end{figure}

RoboAtom-Data provides unified atomic-action annotations for 1.45 million robot trajectories from 16 sources. Figure~\ref{fig:pipeline} summarizes its construction: data collection, human-in-the-loop seed annotation, same-task annotation extension with PhysExtend, and quality control. We describe these stages below and analyze the resulting data in Section~\ref{sec:data-statistics}.

\subsection{Data Definition and Collection}
\label{sec:definition}

Each atomic action is annotated as a segment with three fields: a skill category, an action description, and start and end frames (Figure~\ref{fig:overview}, center). The skill category is a label shared across sources, the action description retains the manipulated object and action details, and the start and end frames align both with the trajectory. A new segment starts only when the dominant skill changes; preparation, arm repositioning, and transitions are absorbed into the adjacent segment (Appendix~\ref{app:pipeline}).

The released set retains the full 29-class skill vocabulary; semantically similar actions are grouped into the same category, with the specific meaning kept in the action description. The full vocabulary is given in Appendix~\ref{app:vocabulary}.

We collect real-robot and high-fidelity simulated data from public datasets and internal recordings, retain tasks with multiple repeated demonstrations after screening by embodied-AI practitioners, and convert all data to the LeRobot v2.1 format~\citep{lerobot}. Table~\ref{tab:datasets} compares RoboAtom-Data with existing subtask-annotated robot datasets; Appendix~\ref{app:roboatom-data} lists all sources and their counting rules.

We reuse available segment annotations from AgiBot World (Beta and 2026), Galaxea Open-World, RH20T, and RoboCerebra, and standardize their skill semantics and action granularity. We first use a locally deployed Qwen3.8-27B to screen source labels and action descriptions, applying source-specific rules where mappings to the unified vocabulary are unambiguous. Composite annotations are decomposed using source-specific rules and available temporal evidence. Uncertain cases undergo visual review with Qwen3.8-27B and targeted human spot checks to identify recurring issues and guide rule-based corrections. Cases unresolved by these steps undergo further visual review with GPT-5.6-sol to assess action semantics and segmentation granularity. For example, some RH20T annotations combine multiple pick-and-place cycles involving different colored blocks into a single segment, requiring finer segmentation to distinguish the constituent atomic actions. Remaining uncertain cases are referred to human annotators for final review and correction. Source-specific mapping rules are detailed in Appendix~\ref{app:vocabulary}.

\begin{table}[!th]
    \centering
    \caption{Robot datasets with subtask annotations. $\dagger$ denotes statistics recomputed by us; source versions and counting rules are in Appendix~\ref{app:roboatom-data}.}
    \label{tab:datasets}
    \small
    \setlength{\tabcolsep}{3pt}
    \begin{tabular}{llrrrrrr}
        \toprule
        Dataset & Domain & Hours & Trajectories & Segments & Skills & Tasks & Embod. \\
        \midrule
        AgiBot World Beta & Real & 2,976.4 & 141,563$^\dagger$ & 1,001,552 & 77$^\dagger$ & 217 & 1 \\
        AgiBot World 2026 & Real & 591.6$^\dagger$ & 28,331$^\dagger$ & 327,355$^\dagger$ & 43$^\dagger$ & 37$^\dagger$ & 1$^\dagger$ \\
        Galaxea Open-World & Real & 500.0 & 19,183$^\dagger$ & 108,775$^\dagger$ & 58 & 150 & 1 \\
        RoboInter-Data & Real & 2,417.1$^\dagger$ & 235,880 & 752,388 & 15 & 232 & 6 \\
        BEHAVIOR-1K 2026 & Sim & 1,952.9 & 20,000 & 406,341 & 35 & 100 & 1 \\
        \textbf{RoboAtom-Data} & \textbf{Mixed} & \textbf{16,882.1$^\dagger$} & \textbf{1,446,974$^\dagger$} & \textbf{6,876,282$^\dagger$} & \textbf{29} & \textbf{3,334$^\dagger$} & \textbf{32$^\dagger$} \\
        \bottomrule
    \end{tabular}
\vspace{-0.2cm}
\end{table}

\FloatBarrier

\subsection{Human-in-the-Loop Annotation}
\label{sec:seed}

Human annotators establish the action semantics and temporal boundaries of the seed annotations used by PhysExtend. Gemini 3.6 Flash assists by generating ordered candidate segments and initial boundaries. Consistent with the findings of \citet{wgo}, we find that contact-sheet inputs substantially reduce annotation costs while maintaining annotation quality. We therefore use a fixed $3\times4$ grid of sampled video frames for subsequent model-assisted annotation and reference-model evaluation. Annotators review the original videos, correct skill categories and action descriptions, adjust segmentation granularity and step order, and refine the start and end frames of each action.

Qwen3-VL-235B-A22B-Instruct~\citep{qwen3vl} then flags spelling errors and skill--description inconsistencies and proposes revisions. Annotators assess these suggestions against the original videos and make the final corrections. The resulting human-reviewed annotations serve as references for same-task extension. The human annotation and refinement cost is approximately 100 USD per source-video hour. Annotation instructions, interfaces, and cleaning rules are provided in Appendix~\ref{app:data}.

\subsection{PhysExtend: Same-Task Annotation Extension}
\label{sec:extension}

Repeated executions of a task often share an ordered skill sequence, allowing boundaries from a few reference trajectories to be propagated. PhysExtend localizes these boundaries in each target trajectory.

For a target trajectory of length $T$ with a known ordered skill sequence $y_{1:M}$, we predict internal boundaries that satisfy the following constraint:
\begin{equation}
    0=b_0<b_1<\cdots<b_{M-1}<b_M=T
    \label{eq:extension-boundary-constraints}
\end{equation}

The problem is modeled as global sequence partitioning with order constraints:
\begin{equation}
    \begin{aligned}
        \hat{\mathbf b}=\arg\min_{\mathbf b}\; &\sum_{m=1}^{M}\left[C_m(b_{m-1},b_m)+\lambda_dD_m(b_{m-1},b_m)\right] \\
        &+\lambda_q\sum_{m=1}^{M-1}Q_m(b_m)
    \end{aligned}
    \label{eq:extension-objective}
\end{equation}

\begin{figure}[t]
    \centering
    \includegraphics[width=0.85\linewidth]{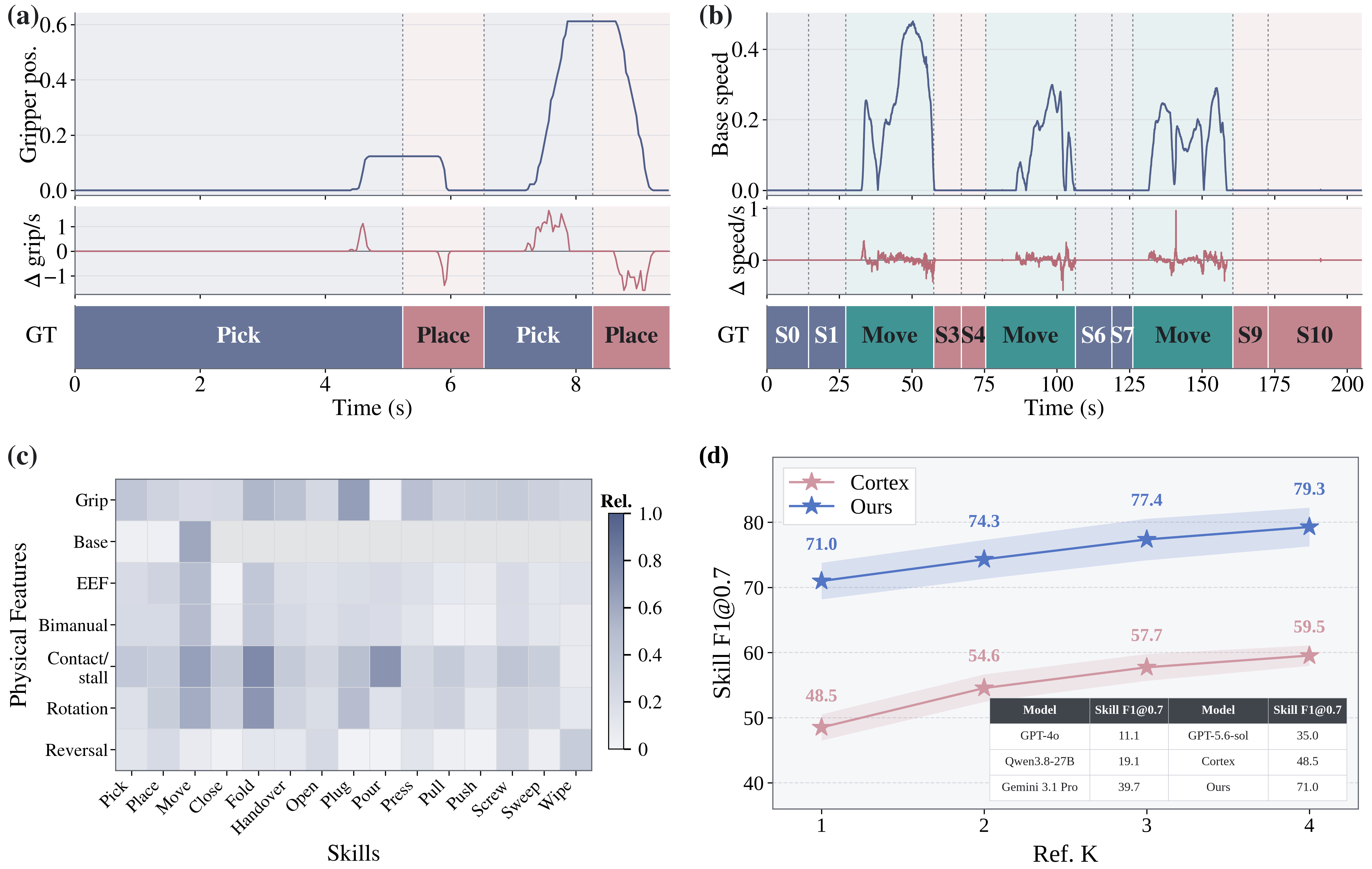}
    \caption{Skill-aware boundary evidence and extension performance. (a) Gripper-state signals aligned with Pick/Place boundaries. (b) Base-motion signals aligned with Move boundaries. (c) Skill-feature associations. (d) PhysExtend consistently outperforms Cortex across reference counts; shaded bands indicate $\pm$1 standard deviation.}
    \label{fig:physical-evidence}
\vspace{-0.2cm}
\end{figure}

Here, $C_m$, $D_m$, and $Q_m$ are penalty terms for deviation from the reference position and motion progress, for segment duration, and for skill-dependent gripper events, respectively. PhysExtend selects ordered boundaries jointly by dynamic programming and then applies physical-event correction and monotonic projection. Its physical evidence is derived from state-action signals, including motion intensity, pauses, direction changes, gripper state, and holding-state consistency. For multiple reference trajectories, each reference path is decoded independently and corresponding boundaries are fused before monotonic projection, improving stability while preserving the shared skill sequence (Figure~\ref{fig:physical-evidence}). Cortex~\citep{cortex} instead fits reference feature distributions and can miss skill transitions when speed, pauses, or retries change (Figure~\ref{fig:boundary-comparison}). Implementation details are given in Appendix~\ref{app:rule-extension}.

The extension evaluation contains 100 tasks, each with 5 manually annotated and reviewed trajectories, and the ordered skill sequence is identical within each task. Given $k$ of them as references, the remaining $5-k$ trajectories are segmented. Skill F1@0.7 requires that the skill labels agree and that the temporal intersection over union is at least 0.7.

\begin{table}[t]
    \centering
    \caption{Annotation extension performance and cost. Episode accuracy (EA) requires all subtasks in a trajectory to have correct semantics and segmentation, evaluated on 100 trajectories, one per task. Cost is in USD per source-video hour.}
    \label{tab:extension}
    \small
    \setlength{\tabcolsep}{2pt}
    \begin{tabular}{lccccc}
        \toprule
        Model & mIoU $\uparrow$ & F1@0.7 $\uparrow$ & Skill F1@0.7 $\uparrow$ & EA (\%) $\uparrow$ & Cost $\downarrow$ \\
        \midrule
        GPT-4o & 48.2 & 20.7 & 11.1 & 10 & 0.888 \\
        Gemini 3.1 Pro & 61.1 & 44.9 & 39.7 & 39 & 2.877 \\
        GPT-5.6-sol & 58.9 & 37.8 & 35.0 & 46 & 3.510 \\
        \midrule
        Qwen3.8-27B & 46.1 & 25.4 & 19.1 & 25 & 0.051 \\
        Cortex & 66.5 & 49.8 & 48.5 & 37 & $4.8\times10^{-4}$ \\
        \textbf{PhysExtend} & \textbf{76.8} & \textbf{71.6} & \textbf{71.0} & \textbf{69} & $\mathbf{2.8\times10^{-5}}$ \\
        \bottomrule
    \end{tabular}
\vspace{-0.5cm}
\end{table}

PhysExtend achieves 71.0 Skill F1@0.7 with one reference trajectory, compared with 48.5 for Cortex, and improves to 79.3 with four references (Figure~\ref{fig:physical-evidence}(d)). Using only state-action signals, PhysExtend processes 15,952 trajectories per CPU hour at a compute cost of $2.8\times10^{-5}$ USD per source-video hour (Appendix~\ref{app:rule-extension}).

Temporal overlap alone does not fully capture annotation quality: long segments can retain high IoU despite substantial boundary offsets, and some transitions admit multiple reasonable boundary placements. We therefore complement automatic metrics with human review, counting an episode as correct only when all subtasks have correct semantics and event-aligned segmentation. GPT-5.6-sol achieves 46\% episode accuracy, a strong result under this strict criterion. PhysExtend reaches 69\% episode accuracy, compared with 37\% for Cortex, closely matching its automatic Skill F1@0.7 score of 71.0. These results suggest that skill-aware physical evidence helps localize skill transitions and align predicted boundaries with physical action events (Table~\ref{tab:extension}).

\subsection{Quality Control}
\label{sec:quality}

We conduct manual quality checks by sampling 20 trajectories from each dataset--embodiment group, examining the correctness of action sequences and temporal boundaries, the consistency of action descriptions with the target trajectories, systematic errors in reference annotations, and annotation consistency across seeds of the same task. Based on the findings, we improve reference-annotation quality, refine the annotation propagation method, and apply a multi-stage review process to correct semantic labels.

Propagation errors mainly arise from differences in execution structure within a task, sensor representations across embodiments, and reference annotations. First, different executions of the same task may contain different operation steps. For example, in a sweeping task, some trajectories continue by dumping the collected trash into a bin, whereas others end after sweeping. Since the reference and target trajectories no longer share the same skill sequence, we exclude such tasks entirely from annotation extension. Second, sensor signals do not correspond to physical skills in the same way across datasets and embodiments, motivating targeted improvements to temporal alignment. For Agilex bimanual robots in the RoboMind datasets, we transfer reference boundaries through dynamic time warping of the full joint trajectories to accommodate differences in execution speed. For mobile bimanual robots, we additionally combine joint, gripper, and chassis states to identify base-motion phases that arm signals alone cannot distinguish. Where action boundaries exhibit stable offsets from grasp and release events, we calibrate these offsets using same-task references. We also improve reference-sequence selection by jointly considering motion similarity, gripper-event counts, and execution duration, and restrict fusion between semantically incompatible references to reduce errors caused by differences in action counts, execution order, and annotation granularity.

Finally, reference annotations are the starting point of PhysExtend, and their quality directly affects the accuracy of propagated annotations. The main issues are errors in the original annotations and inconsistencies across seeds of the same task. Incorrect skill categories, action descriptions, or temporal boundaries in a reference can propagate to multiple trajectories of the same task, producing systematic errors. We identify and correct these errors through downstream quality checks and cross-checks among seeds. Inconsistencies across seeds can also disrupt correspondence between reference segments and cause errors in boundary fusion. In our quality checks, the subtask segmentation accuracy and episode accuracy are 87.6\% and 80.5\%, respectively, when seed sequences are consistent, compared with 79.5\% and 57.4\% when they are inconsistent. On some simulation benchmarks, subtask segmentation accuracy even reached 100\% in the sampled trajectories. Such inconsistencies can reflect either genuine differences in execution steps or differences in annotators' interpretations of the same execution structure. We conduct a second manual quality check of annotations with inconsistencies across seeds; for reference trajectories with the same execution structure, we standardize skill sequences, segmentation granularity, and boundary criteria. For tasks with insufficient reference quality or limited coverage of execution variants, we also manually annotate additional reference trajectories.

Even when reference and target trajectories share the same skill sequence, their manipulated objects, colors, or execution order may differ, so action descriptions cannot be copied directly from the references. For example, a reference trajectory may pick up an apple before an orange, while the target follows the reverse order; a block-manipulation trajectory may retain the same skill sequence but involve blocks of different colors. We first use a locally deployed Qwen3.8-27B to visually check annotations against task instructions and multi-view images. GPT-5.6-sol then reviews and corrects them using same-task reference annotations, the unified skill vocabulary, and field constraints, followed by final human verification. Revisions preserve the established action granularity and language style, avoiding unsupported rewrites of reasonable descriptions, such as changing ``turn on the stove'' to ``rotate the knob'' or merging Pick and Place into Move. Semantic correction preserves the original temporal segmentation unless visual evidence reveals an obvious missing segment, extra segment, or severe boundary offset, in which case humans adjust the segmentation.

We require a minimum subtask segmentation accuracy of 85\% and episode accuracy of 75\% for release. We will iteratively correct annotations and reassess their quality, releasing the data only after both criteria are met.

\subsection{Data Statistics and Analysis}
\label{sec:data-statistics}

\begin{figure}[t]
    \centering
    \includegraphics[width=0.9\linewidth]{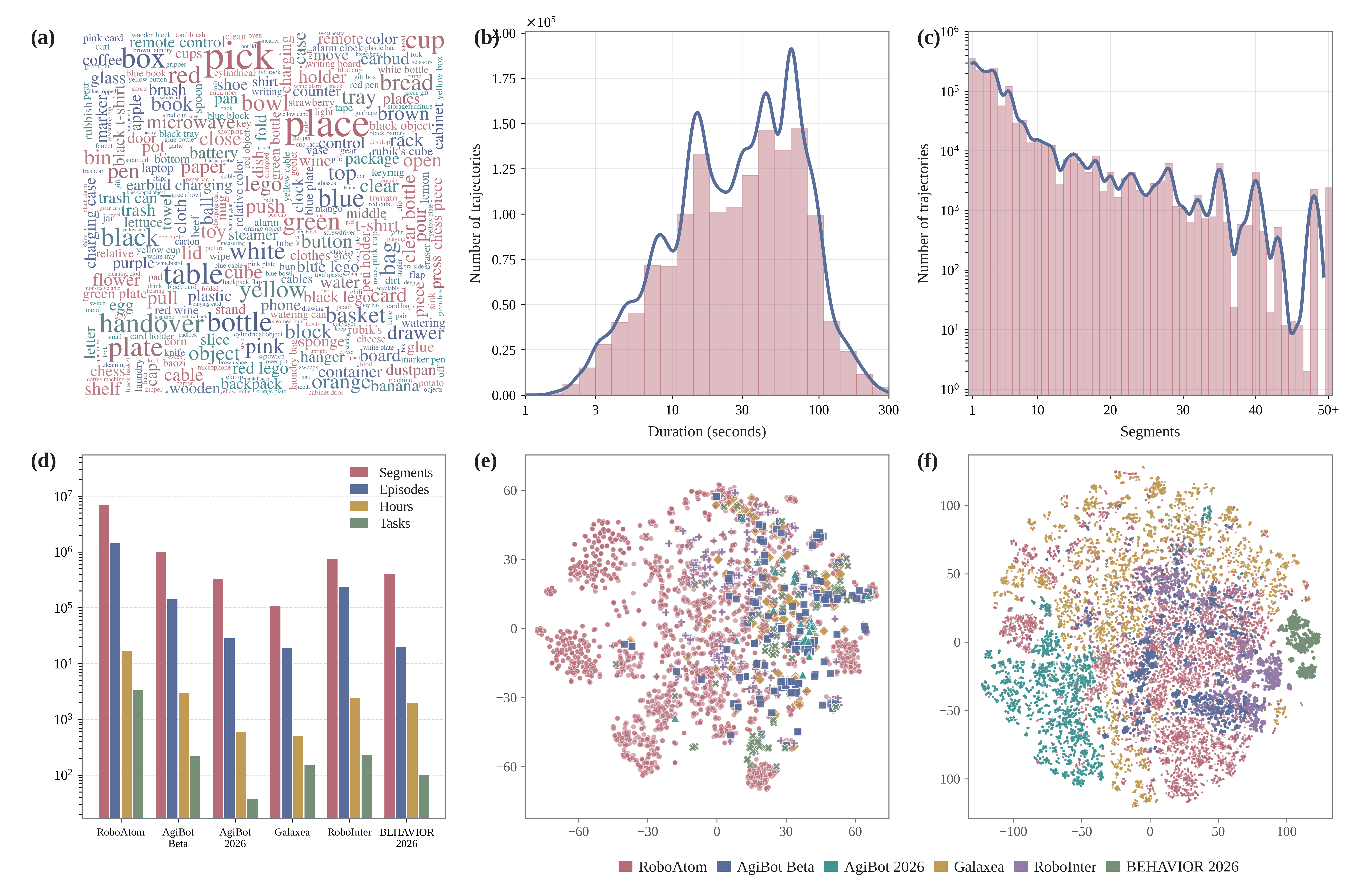}
    \caption{Scale and diversity of RoboAtom-Data. (a) Annotation word cloud. (b) Trajectory durations. (c) Segments per trajectory. (d) Dataset-scale comparison. (e--f) Task-instruction and action-description distributions.}
    \label{fig:data-characteristics}
    \vspace{-0.5cm}
\end{figure}

RoboAtom-Data contains 1,446,974 trajectories, 6,876,282 skill-labeled segments, and 16,882.1 hours across 16 sources and 3,334 tasks, including approximately 9,375 real-robot hours and 7,508 simulated hours. It exceeds all compared subtask-annotated datasets in trajectories, segments, hours, and tasks (Figure~\ref{fig:data-characteristics}(d); Table~\ref{tab:datasets}).

Figures~\ref{fig:data-characteristics}(a--c) illustrate diversity in manipulation content and task length. The relatively similar word sizes in the annotation word cloud suggest balanced coverage of manipulated objects and atomic actions. Trajectory durations range from a few seconds to several minutes, and the number of action segments per trajectory ranges from one to more than 50, covering different skills and manipulated objects as well as tasks ranging from short interactions to long-horizon, multi-step operations.

Beyond scale, Figures~\ref{fig:data-characteristics}(e--f) qualitatively illustrate diversity at both the task and atomic-action levels. RoboAtom-Data spans broad regions of the task-instruction and action-description projections, covering varied task goals as well as object-specific operations. Overlap with other datasets indicates shared manipulation semantics, while regions occupied by different sources illustrate complementary language coverage. Together, these views highlight the breadth of supervision obtained by integrating heterogeneous robot datasets. Per-source statistics and dataset-specific language visualizations are provided in Appendices~\ref{app:roboatom-data} and~\ref{app:language-space}.

\section{RoboAtom-Bench}
\label{sec:benchmark}

RoboAtom-Bench evaluates whether a model can recover atomic-action annotations from robot video: which skills occur, in what order, and where each one begins and ends. Annotation difficulty grows with the number of actions in a trajectory and changes when a model must combine several camera views, so the benchmark separates these conditions into subsets with a shared vocabulary and matching protocol.

\begin{figure}[t]
    \centering
    \includegraphics[width=0.9\linewidth]{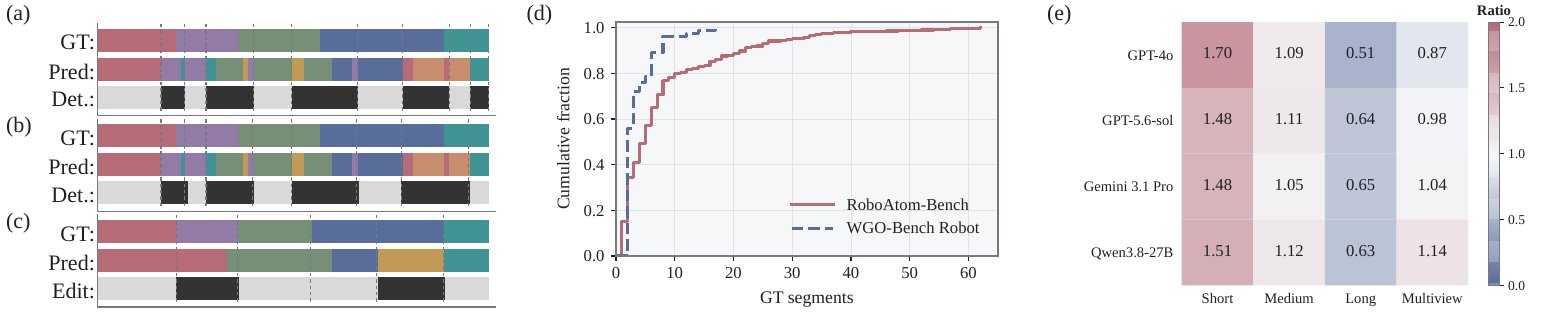}
    \caption{(a--c) Illustrations of temporal mIoU/F1, Skill-F1, and Edit. Dark regions indicate unmatched predictions or edit operations. (d) Cumulative distributions of GT segment counts. (e) Ratios of total predicted to total GT segment counts across RoboAtom-Bench subsets.}
    \label{fig:benchmark-analysis}
\end{figure}

\subsection{Benchmark Construction and Composition}
\label{sec:benchmark-construction}

Given a video and task instruction, a model predicts ordered segments with temporal boundaries, skill categories, and action descriptions. From the human-annotated data, we screen 800 candidate trajectories by data source, view information, and number of segments. Embodied-data experts review their semantics, granularity, and boundaries, retaining 500 trajectories with 4,018 ground-truth (GT) segments, which are supplemented with view annotations. The benchmark uses 15 skills covering 99.2\% of RoboAtom-Data skill-labeled segments (Appendix~\ref{app:vocabulary}).

Short, Medium, and Long contain 200, 100, and 100 trajectories with 1--4, 5--8, and 9--62 GT actions, respectively, and examine annotation performance as sequence length grows. Multiview contains 100 separate trajectories with synchronized camera inputs. Figure~\ref{fig:benchmark-analysis}(d) shows the broader action-count distribution of RoboAtom-Bench, with a median of five segments versus two for WGO-Bench Robot (Appendix~\ref{app:test-distribution}).

\subsection{Evaluation Metrics}
\label{sec:benchmark-metrics}

An annotation can fail in three ways: its boundaries can be misplaced, its skills can be wrong, or its action sequence can have missing, extra, or reordered steps. We therefore report four complementary metrics on RoboAtom-Bench (Figure~\ref{fig:benchmark-analysis}(a--c)). Temporal mIoU and temporal F1 evaluate segment overlap regardless of skill labels (panel (a)); Skill F1 requires both sufficient temporal overlap and matching skill labels (panel (b)); and Edit Score compares skill sequences without using timestamps (panel (c)). For GT interval $G_i$ and prediction $P_j$, we compute temporal overlap and mIoU as follows:
\begin{equation}
    \operatorname{tIoU}(G_i,P_j)=\frac{|G_i\cap P_j|}{|G_i\cup P_j|},
    \qquad
    \operatorname{mIoU}=\frac{1}{N_G}\sum_{i=1}^{N_G}\operatorname{tIoU}(G_i,P_{\pi(i)})
    \label{eq:benchmark-miou}
\end{equation}
Here, $N_G$ is the number of GT segments pooled over all episodes and $\pi$ is an order-preserving one-to-one alignment maximizing total IoU, independently of skill labels; unmatched GT segments contribute zero. Both F1 metrics pool $TP$, $FP$, and $FN$ across episodes, then compute the score as follows:
\begin{equation}
    F1=\frac{2TP}{2TP+FP+FN}
    \label{eq:benchmark-f1}
\end{equation}
Each F1 metric builds its own matches. Temporal F1 uses an order-preserving one-to-one alignment that maximizes the number of pairs with IoU at least threshold $\tau$, regardless of skill. Skill F1 follows the segmental F1 of \citet{lea2017tcn} as implemented in MS-TCN~\citep{mstcn}: predictions are processed in temporal order, and each is matched to its highest-IoU GT segment of the same skill if that IoU is at least $\tau$ and the GT segment is not already matched. Unmatched predictions and GT segments are false positives and false negatives. The main results use $\tau=0.7$.

Edit compares skill sequences $S_P$ and $S_G$ after collapsing adjacent identical labels:
\begin{equation}
    \operatorname{Edit}=100\left(1-\frac{d_{\mathrm{Lev}}(S_P,S_G)}{\max(|S_P|,|S_G|)}\right)
    \label{eq:benchmark-edit}
\end{equation}
Here, $d_{\mathrm{Lev}}$ is the Levenshtein distance, following the segmental edit score of \citet{lea2017tcn}. Edit measures skill insertions, deletions, and substitutions independently of temporal boundaries and is averaged over episodes. mIoU and both F1 scores are calculated on a 0--1 scale and multiplied by 100 when reported; Edit Score is defined directly on a 0--100 scale. Implementation details are in Appendix~\ref{app:eval-metrics}.

\subsection{Benchmark Analysis}
\label{sec:benchmark-analysis}

We analyze four reference models on RoboAtom-Bench: GPT-4o, GPT-5.6-sol, Gemini 3.1 Pro, and Qwen3.8-27B.

Reference models can recover skill order while missing precise action boundaries. At the benchmark level, GPT-5.6-sol, the strongest reference model, scores 63.4 on Edit and 28.0 on temporal F1@0.7. One possible explanation is that video-level descriptions used to train many video LLMs identify what happens without marking when each event begins and ends; such supervision may support semantic recognition while leaving precise temporal segmentation weak~\citep{vtimellm}.

Segmentation granularity also shifts systematically with sequence length (Figure~\ref{fig:benchmark-analysis}(e)). The four analyzed models predict 1.48--1.70 times the GT segment count on Short but only 0.51--0.65 times on Long. Inspected Short cases include extra segments and joint actions split into multiple predictions; Long cases include successive actions merged into one segment. Long trajectories are also longer on average (99 versus 13 seconds for Short; Appendix~\ref{app:test-distribution}) and contain more actions shorter than two seconds (21.8\% versus 11.7\%). At the shared 2-FPS input rate, these brief actions span only a few sampled frames, so missed transitions and timing errors may contribute to merged segments and lower overlap. At the same time, compressing longer videos into limited visual tokens can lose fine temporal detail, which may be another reason for merged predictions on Long~\citep{timechat}.

\section{Learning with Atomic-Action Supervision}
\label{sec:method}
\label{sec:validation}

We use the time-aligned annotations in RoboAtom-Data (Section~\ref{sec:dataset}) as prediction targets for automatic video annotation and as intermediate instructions for robot control.

\subsection{RoboAtom-Anno: Automatic Atomic-Action Annotation}
\label{sec:anno-method}

Starting from Qwen3.5-9B~\citep{qwen35}, we perform robot-domain mid-training with gripper-state question answering, object grounding, spatial reasoning, and action planning from images or frame sequences, using 3,084.1 GPU hours on 48 GPUs. We then fine-tune the model on a task-sampled subset of RoboAtom annotations to generate segment sequences from complete videos for 500 steps, using 27.7 GPU hours on 8 GPUs.

Finally, we optimize annotation quality with GRPO~\citep{grpo}, using 364.5 GPU hours on 16 GPUs. Training videos are stratified by offline prediction quality, emphasizing medium and hard cases while retaining easy examples. For each video, multiple candidate annotations are sampled and scored against the reference with the reward:
\begin{equation}
R=\lambda_{\mathrm{temporal}}R_{\mathrm{temporal}}+\lambda_{\mathrm{soft}}R_{\mathrm{soft}}+\lambda_{\mathrm{edit}}R_{\mathrm{edit}}+\lambda_{\mathrm{format}}R_{\mathrm{format}}
\end{equation}
Here, $R_{\mathrm{temporal}}$ rewards ordered temporal matches, $R_{\mathrm{soft}}$ measures continuous temporal overlap, $R_{\mathrm{edit}}$ measures skill-sequence edit similarity, and $R_{\mathrm{format}}$ checks parsing and field constraints. Data construction is detailed in Appendix~\ref{app:training-data}; training settings and reward weights are in Appendices~\ref{app:mid-training}, \ref{app:sft}, and~\ref{app:grpo}.

\subsection{Dual-System Policy Learning and Execution}
\label{sec:policy-method}

\begin{figure}[t]
\vspace{-0.5cm}
    \centering
    \includegraphics[width=0.75\linewidth]{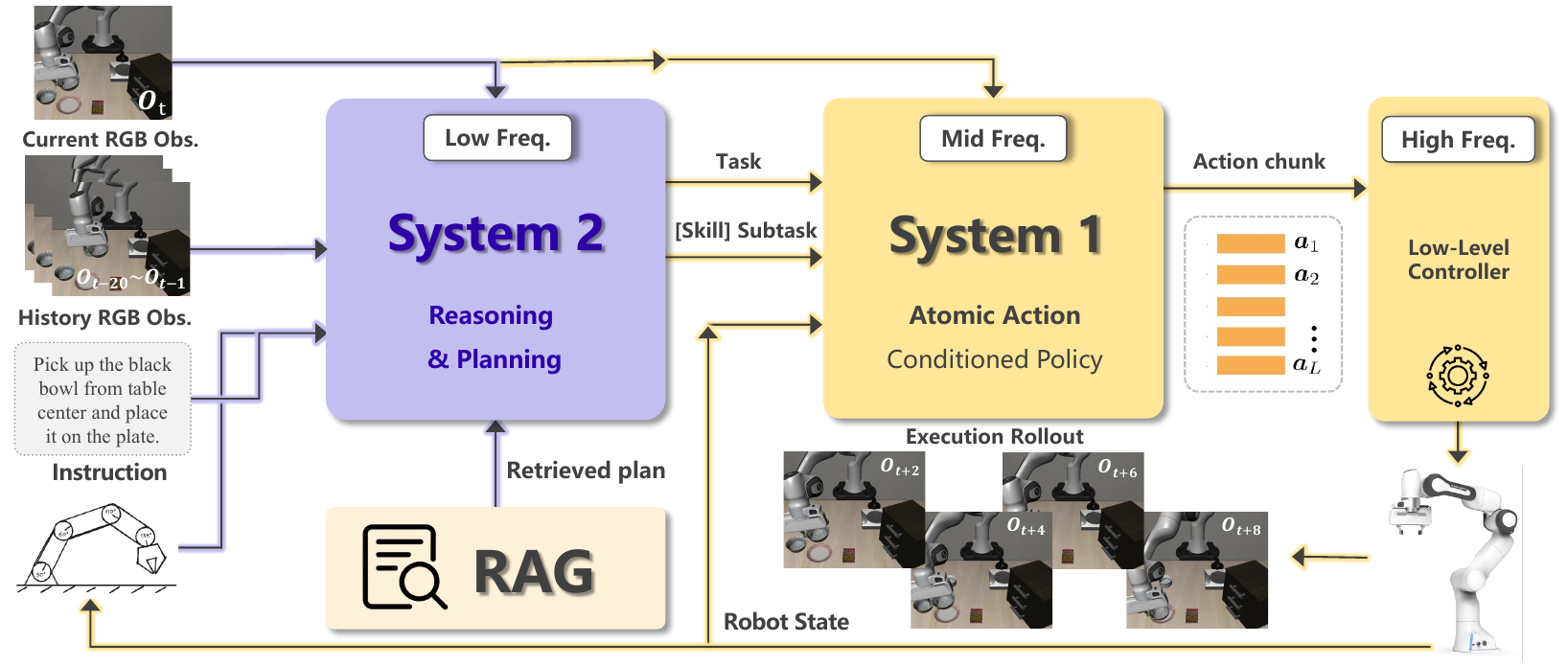}
    \caption{Dual-system execution. System~2 uses a task-plan prior, visual observations, and execution history to select an atomic-action instruction. System~1 conditions $\pi_{0.5}$ on that instruction to produce robot actions; subsequent observations inform the next planning decision.}
    \label{fig:dual-system}
\end{figure}

Inspired by Hi Robot~\citep{hirobot}, we use atomic-action supervision in a dual-system architecture (Figure~\ref{fig:dual-system}). During training, each observation--action pair receives the skill label and action description of its covering segment. Together with the task instruction, these condition the System~1 policy, which we obtain by fine-tuning $\pi_{0.5}$~\citep{pi05} with OpenPI~\citep{openpi}.

At deployment, System~2 supplies the intermediate instructions required by the trained policy. A task-plan prior from the task knowledge base provides candidate skills and action descriptions, which the Qwen3.8-27B planner combines with visual observations and execution history to select the current instruction. System~1 executes the selected instruction through action chunks, and the resulting observations guide System~2 in retaining the current instruction or switching to another atomic action. Implementation details are in Appendix~\ref{app:executor}.

\subsection{Annotation Experiments}
\label{sec:anno}

We compare RoboAtom-Anno with the four reference models of Section~\ref{sec:benchmark-analysis} and the base Qwen3.5-9B on RoboAtom-Bench and WGO-Bench Robot. RoboAtom-Bench tasks are excluded from training (Appendix~\ref{app:training-data}). On WGO-Bench Robot, we report mIoU and temporal F1@0.7 under its open-vocabulary annotations. Evaluation settings are given in Appendix~\ref{app:eval-metrics}.

The RoboAtom-Anno panel of Figure~\ref{fig:overview} traces Skill F1@0.7 across training configurations. From 3.5 for the base model, SFT on noisy open-source annotations reaches 6.6, and vocabulary alignment with ambiguity filtering raises it to 17.1 (Appendix~\ref{app:training-data}). Replacing these with RoboAtom supervision and training language-only LoRA reaches 23.0, comparable to GPT-5.6-sol (21.4). Adopting the best-performing SFT configuration further raises Skill F1@0.7 to 24.2 (Appendix~\ref{app:visual-update}).

Initializing from robot-domain mid-training and applying the same SFT configuration raises Skill F1@0.7 from 24.2 to 26.9 and mIoU from 52.9 to 54.6 (Appendix~\ref{app:training-config}). Continuing with RLVR using GRPO raises Skill F1@0.7 to 30.1 and temporal F1@0.7 from 35.5 to 41.0, while the predicted-to-GT segment-count ratio decreases from 1.20 to 0.99.

\begin{table}[!htbp]
    \centering
    \caption{Annotation performance and inference cost (USD per source-video hour; Appendix~\ref{app:gpu-cost}).}
    \label{tab:benchmark-reference}
    \small
    \setlength{\tabcolsep}{3pt}
    \begin{tabular}{lcccccccc}
        \toprule
        \multirow{2}{*}{Model} & \multicolumn{5}{c}{RoboAtom-Bench} & \multicolumn{3}{c}{WGO-Bench Robot} \\
        \cmidrule(lr){2-6} \cmidrule(lr){7-9}
         & mIoU $\uparrow$ & F1@0.7 $\uparrow$ & Skill F1@0.7 $\uparrow$ & Edit $\uparrow$ & Cost $\downarrow$ & mIoU $\uparrow$ & F1@0.7 $\uparrow$ & Cost $\downarrow$ \\
        \midrule
        GPT-4o & 31.5 & 13.9 & 7.2 & 47.7 & 0.971 & 46.1 & 14.6 & 0.800 \\
        GPT-5.6-sol & 44.0 & 28.0 & 21.4 & 63.4 & 4.330 & 41.9 & 31.3 & 3.375 \\
        Gemini 3.1 Pro & 42.5 & 24.4 & 16.1 & 61.8 & 4.516 & 52.1 & 29.6 & 4.389 \\
        Qwen3.8-27B & 38.4 & 20.1 & 10.9 & 56.6 & 0.103 & 51.3 & 24.3 & 0.096 \\
        Qwen3.5-9B & 9.6 & 6.2 & 3.5 & 28.7 & 0.008 & 12.8 & 1.7 & 0.007 \\
        \midrule
        RoboAtom-Anno-9B & 55.1 & 41.0 & 30.1 & 76.7 & \textbf{0.006} & 61.7 & 48.3 & \textbf{0.004} \\
        RoboAtom-Anno-27B & \textbf{56.4} & \textbf{43.2} & \textbf{33.8} & \textbf{76.8} & 0.078 & \textbf{66.0} & \textbf{49.2} & 0.054 \\
        \bottomrule
    \end{tabular}
\end{table}

The final RoboAtom-Anno-9B outperforms all reference models on every reported quality metric at the lowest inference cost (Table~\ref{tab:benchmark-reference}). On RoboAtom-Bench, it reaches 30.1 Skill F1@0.7 versus 21.4 for GPT-5.6-sol, at 0.006 versus 4.330 USD per video hour (Appendix~\ref{app:gpu-cost}). Results by RoboAtom-Bench subset are provided in Appendix~\ref{app:subset-difficulty}.

On WGO-Bench Robot, RoboAtom-Anno-9B reaches 48.3 temporal F1@0.7 versus 31.3 for the strongest reference model, a gain of 17.0 points. On WGO-Bench Ego, it reaches 21.1 versus the best reference result of 14.1.

To explore whether our training recipe benefits from a stronger pretrained backbone, we also train RoboAtom-Anno-27B from Qwen3.8-27B using the same recipe. It achieves 33.8 Skill F1@0.7 on RoboAtom-Bench and 49.2 temporal F1@0.7 on WGO-Bench Robot, improving over the 9B variant by 3.7 and 0.9 points, respectively. On WGO-Bench Ego, temporal F1@0.7 increases from 21.1 to 28.3. These results suggest that a stronger pretrained backbone can further improve annotation quality and transfer to egocentric human manipulation videos.

\subsection{Robot Control Experiments}
\label{sec:policy}

We evaluate long-horizon control on LIBERO-Long~\citep{libero}, robustness to perturbations on LIBERO-Plus~\citep{liberoplus}, and bimanual manipulation under Clean and Random conditions on RoboTwin~\citep{robotwin}. We compare task-level post-training and end-to-end execution of $\pi_{0.5}$ with atomic-action-conditioned post-training and System~2 scheduling at inference. Both settings use the same post-training budget, observation space, and action space within each benchmark. Image augmentation reduced baseline performance on LIBERO-Long and RoboTwin Clean in our control experiments. We therefore train both methods without image augmentation (Appendices~\ref{app:libero-aug}--\ref{app:robotwin-aug}).

\begin{table}[H]
    \centering
    \caption{Downstream success rates (\%) on LIBERO, LIBERO-Plus, and RoboTwin. LIBERO-Plus Overall is the success rate over all 10,030 tasks; RoboTwin Avg. is the Clean/Random mean.}
    \label{tab:downstream}
    \small
    \setlength{\tabcolsep}{4pt}
    \begin{tabular}{lccccccccc}
        \toprule
        \multirow{2}{*}{Method} & LIBERO & \multicolumn{5}{c}{LIBERO-Plus} & \multicolumn{3}{c}{RoboTwin} \\
        \cmidrule(lr){2-2} \cmidrule(lr){3-7} \cmidrule(lr){8-10}
         & Long $\uparrow$ & Spatial $\uparrow$ & Object $\uparrow$ & Goal $\uparrow$ & Long $\uparrow$ & Overall $\uparrow$ & Clean $\uparrow$ & Random $\uparrow$ & Avg. $\uparrow$ \\
        \midrule
        $\pi_{0.5}$ & 95.00 & {\textbf{93.01}} & 85.23 & 79.93 & 78.56 & 84.05 & 80.70 & 48.04 & 64.37 \\
        Our method & {\textbf{98.20}} & 92.46 & {\textbf{91.86}} & {\textbf{86.34}} & {\textbf{82.29}} & {\textbf{88.18}} & {\textbf{82.02}} & {\textbf{50.32}} & {\textbf{66.17}} \\
        \bottomrule
    \end{tabular}
\end{table}

Table~\ref{tab:downstream} shows that our method improves LIBERO-Long success by 3.20 percentage points to 98.20\%, supporting the benefit of atomic-action-conditioned training and hierarchical execution for long-horizon tasks. On LIBERO-Plus, overall success increases by 4.13 points, with gains of 6.63, 6.41, and 3.73 points on Object, Goal, and Long, respectively. These improvements suggest better adaptation to changes in objects and goals, alongside more reliable long-horizon execution. On RoboTwin, success improves by 1.32 points under Clean and 2.28 points under Random, raising the average by 1.80 points to 66.17\%. The gains across both settings extend the benefits to bimanual manipulation.

\section{Limitations}
\label{sec:limitations}

Our evaluation is limited in training scope and data coverage. We have not yet verified the role of atomic-action supervision in large-scale policy pretraining. Scaling real-robot data collection also requires substantial hardware and human effort, while egocentric human video, whose volume far exceeds that of robot data, has not yet been included. Extending to this source poses a temporal-granularity challenge: human atomic actions are often brief and fragmented, with some lasting less than one second, making their boundaries difficult to localize at the current sampling rate. Robot supervision provides a useful starting point, but adapting the approach to human video may require a coarser definition of atomic actions.

\section{Conclusion}
\label{sec:conclusion}

We presented RoboAtom, which addresses missing and inconsistent subtask annotations across heterogeneous robot datasets. RoboAtom-Data unifies action semantics and temporal boundaries through a shared annotation protocol, human annotation, same-task propagation with PhysExtend, and multi-stage quality assessment and correction. RoboAtom-Bench provides skill-aware temporal evaluation of atomic-action annotation. Experiments demonstrate the resulting data's value in two applications: RoboAtom-Anno achieves better automatic annotation performance than the evaluated open- and closed-source models, while atomic-action supervision paired with hierarchical execution improves robot task success over task-level baselines.

\FloatBarrier

\bibliography{roboatom}
\bibliographystyle{roboatom}

\restorefloatlimits
\appendix
% Number appendix figures and tables by section (A.1, B.1, ...)
\counterwithin{figure}{section}
\counterwithin{table}{section}
\renewcommand{\theHfigure}{\Alph{section}.\arabic{figure}}
\renewcommand{\theHtable}{\Alph{section}.\arabic{table}}

\section{Data and Annotation Specification}
\label{app:data}

\subsection{Human and Model Collaborative Annotation Pipeline}
\label{app:pipeline}

Pre-annotation is produced by Gemini 3.6 Flash, which generates atomic-action segments and initial boundaries from the video and the task description; the prompt used is given below.

\begin{lstlisting}
Analyze the robotic-manipulation video and return atomic action segments.

Input:
- sample_id: ${sample_id}
- total_frames: ${total_frames}
- fps: ${fps}
- task_instruction: ${task_instruction}
- allowed_skills: [Pick, Place, Handover, Move, Open, Close, Push, Pull, Press, Rotate, Screw, Plug, Pour, Fill, Scoop, Stir, Dip, Shake, Fold, Tie, Wipe, Sweep, Wash, Brush, Cut, Peel, Strike, Scan]

Use the frame IDs shown in the images. Frame ranges are closed and inclusive.

Rules:
1. Cover every frame from 0 through ${last_frame} exactly once. Segments must be ordered and contiguous: each start_frame equals the previous end_frame + 1.
2. Split only when the dominant atomic skill changes. Do not split tiny transitions.
3. Absorb preparation, arm repositioning, transitions, and idle frames into the nearest meaningful action. Never output Static.
4. Emit Pick when grasp changes from not holding to holding, and Place when a held object is released. Keep them separate from subsequent actions; for example: Pick tool -> Sweep/Wipe -> Place tool.
5. Move means physical robot-base/chassis translation only. Arm reaching, lifting, lowering, alignment, and retraction belong to the neighboring manipulation.
6. In bimanual scenes, label the dominant active manipulation rather than passive holding or stabilization. Use Handover only for a direct transfer between hands/arms.
7. Use Pull/Push for sliding drawers. Use Open/Close for hinged or articulated doors, lids, covers, caps, or flaps. Treat one continuous opening or closing motion as one segment.
8. Each segment has exactly one allowed skill. Decompose composite behaviors. If the video has one meaningful action, return one segment covering all frames.

Examples:
- Pick: "picks up the sponge" -> Wipe: "wipes the table" -> Place: "places the sponge on the tray".
- Pull: "pulls open the drawer"; Push: "pushes the drawer closed".
- Do not create segments such as "approaches the sponge", "moves the arm", or "remains idle".

Return valid JSON only, with no markdown or extra text:
{
  "sample_id": "${sample_id}",
  "nframes": ${total_frames},
  "fps": ${fps},
  "task": "Overall task description in English",
  "segments": [
    {
      "seg_id": 0,
      "start_frame": 0,
      "end_frame": ${last_frame},
      "subtask": "short English action",
      "primitive_skill": "One allowed atomic skill"
    }
  ]
}

primitive_skill must exactly match an allowed skill, and seg_id must be sequential from 0.
\end{lstlisting}

\begin{figure}[h]
    \centering
    \includegraphics[width=0.96\linewidth]{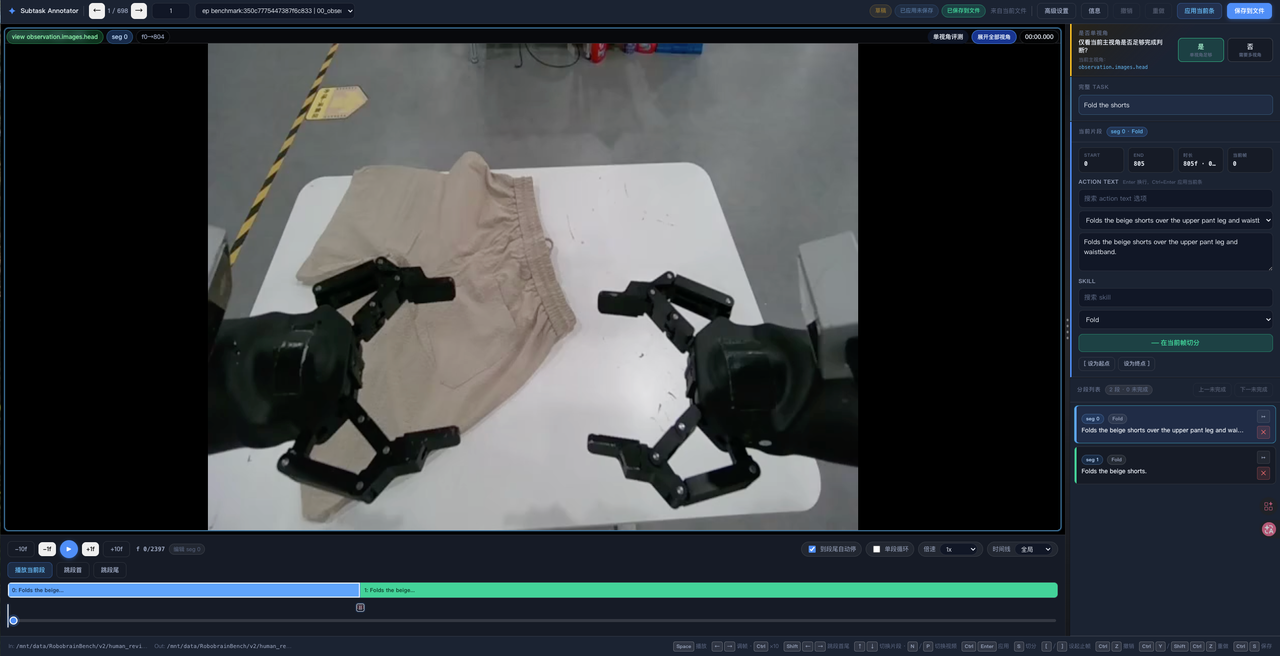}
    \caption{Atomic-action annotation interface.}
    \label{fig:annotation-interface}
\end{figure}

For human annotation and review, annotators correct the action descriptions, categories, and boundaries after pre-annotation in the interface shown in Figure~\ref{fig:annotation-interface}.

After manual correction, Qwen3-VL-235B-A22B-Instruct proposes revisions based on the video and the current annotation; the review interface is shown in Figure~\ref{fig:review-interface}, and the review prompt is given below. Idle and Recover in the prompt are intermediate processing labels: at release, Idle keeps its time interval and text but has an empty skill field and is not counted among valid action segments or skill counts, and Recover is assigned to the corresponding category according to the action actually performed. The released set uses the 29-class vocabulary of Appendix~\ref{app:vocabulary}.

\begin{lstlisting}
You are a strict subtask-local reviewer for human subtask annotations of a robot manipulation video.

Review only annotation quality. Do not grade delivery time, service plan, or aesthetics.
First read the timeline and understand the episode flow, but judge each segment locally.
For a specific segment, the primary evidence is that segment plus its same-object Pick/Place pair.
Do not use later unrelated objects, global bin visibility, or inventory context alone to overturn a local segment.

Input:
- sample_id: <sample_id>
- task: <task_text>
- total_frames: <total_frames>
- fps: <fps>
- review_pass: <review_pass>
- available_view_keys: <available_view_keys_json>
- allowed_skills: <allowed_skills_json>

Human annotation segments use inclusive frame ranges:
<segments_json>

Evidence package:
- The first attached image is a timeline overview with segment ranges, skills, action text, and boundaries.
- The remaining contact-sheet images contain frames sampled strictly every N frames, where N is configured by --frame-stride.
- No segment-internal key frames or boundary-window frames are attached in this mode.
- Use the frame manifest to map each tile to source frame id, segment id, view key, and stride role.

Evidence reading guidance:
Stride-image evidence reading rules:
- Treat the dense stride frames as a frame-by-frame record. Do not decide object identity from a single static tabletop frame.
- If multiple views are available, compare tiles with the same source frame id across views before changing the annotation.
- Spend most of your attention on tiles where the gripper is visible or about to enter/leave the frame. Tiles with no gripper are mostly background context and should not dominate the decision.
- For white/light paper, tissue, wrappers, bags, or cloth, inspect the object touching or held by the gripper very carefully. Small white objects on a dark gripper can be low-contrast; do not miss a white object simply because it is small, partially occluded, or near the gripper edge.
- When judging color, prioritize the object in or immediately under the gripper during grasp, lift, carry, and release. A white object on the gripper is stronger evidence than a similar white object remaining on the table.
- For every Pick/Place pair, first build a manipulated-object trace: before contact -> gripper closure/contact -> object lifted or stably held -> carried object -> release/after-release.
- Identify object color/material from the object inside the gripper during lift, transport, and immediately before release. This gripper-held evidence is stronger than what remains on the table.
- To change a color/material/object name, require held-object evidence from the Pick/Place pair, preferably after lift and before release. A claim based only on which object remains on the table is not enough.
- A table becoming empty is not proof that later Pick/Place segments are hallucinated. Check whether the gripper is already holding an object, whether the object is near the bin, or whether a failed/drop/retry sequence is visible.
- A similar object still visible on the table is not proof that the annotated object was not picked. There may be duplicates, partial occlusion, or a visually similar object; use contact and motion continuity.
- Similar-looking objects are separate instances. If two white/light crumpled papers, tissues, bags, wrappers, or cloth pieces are visible, track the specific instance by its initial position, gripper contact point, carried trajectory, and release location.
- Do not claim a placed object was returned to the table merely because another similar white/light object remains on the table. That remaining object may be a different instance.
- When a Pick/Place pair handles one of multiple similar objects, decide success/failure from whether the held instance leaves the gripper at the annotated target, not from whether a lookalike is still visible elsewhere.
- Do not swap colors between adjacent paper/tissue/plastic objects by inventory reasoning. Only swap object names when the grasped/held object across the local Pick/Place pair clearly has the other color.
- Small crumpled paper, tissue, bags, and wrappers can look different under gripper occlusion, motion blur, or shadows. If the held object's color is not clearly identifiable, keep the original annotation.
- Do not delete or merge retry Pick/Place segments only because the table looks empty after an attempted placement. If the gripper is holding, dropping, re-grasping, or carrying an object, keep the corresponding atomic segments.
- Conversely, do not invent a failed Pick/Place retry from brief hovering, occlusion, dragging, near-table motion, or a momentary pause. A failed attempt requires clear evidence that the gripper opens/releases the held instance, the instance is no longer held, and a later distinct grasp occurs.
- When the contact sheets show conflicting tabletop context and gripper-held evidence, prefer the gripper-held evidence. If the held object is too small, occluded, or out of view, keep the original annotation instead of an incorrect repair.

Segment evidence frame manifest:
<segment_manifest_json>

Global episode evidence frame manifest. Use these only as orientation/background, not as standalone proof of an object/bin error:
<global_manifest_json>

Judge whether the human annotation is correct along two axes, using subtask-local evidence:
1. Description accuracy: skill, action_text, object identity, target/bin/container, and wording.
2. Segmentation accuracy: whether each segment's dominant atomic skill is coherent and adjacent boundaries split only when the dominant atomic skill changes.

Use the same strict error categories internally:
- boundary_error: the boundary cuts a core atomic event incorrectly, merges the core Pick and Place events, puts the grasp/lift/release event in the wrong segment, or lets one segment contain the core event of another subtask.
- wrong_object: action_text names the wrong object, material, color, or manipulated item, proven by clear local evidence.
- wrong_action: action_text or skill names the wrong action, or skill is outside/contradicts the allowed skill set.
- wrong_bin: the destination bin/container/color/location is wrong, proven by the current segment or same-object pair.
- missing_action: an obvious pick/place/action is absent.
- extra_action: a segment describes an action not shown.
- naming_inconsistent: the same physical object is named inconsistently within its Pick/Place pair.
- format_error: text has spelling/format problems that need cleanup.
- unclear_visual: visual evidence is insufficient or occluded after careful inspection.

Segmentation rules:
- Boundaries are important, but do not judge them by duration or by a fixed frame tolerance. Judge whether the dominant atomic skill changed.
- For Pick, the segment should contain the grasp/lift event where the gripper changes from not holding to stably holding the object.
- For Pick, prioritize object continuity at the gripper: compare frames immediately before contact, during gripper closure/lift, and after lift. If the gripper stably holds a visually matching object after the grasp, treat the Pick as successful unless local evidence proves it held a different object.
- Do not mark a Pick failed only because a similar-colored or similar-shaped item remains on the table. There may be duplicate or visually similar objects; the error must be based on the annotated target object not entering or remaining held by the gripper, not just a lookalike still visible elsewhere.
- For Place, the segment should contain the placement/release event where the held object is released and becomes stable at the target.
- Idle is a valid atomic skill at the beginning, middle, or end of an episode when no deliberate task skill is being executed. A human-annotated Idle segment must not be merged into, or relabeled as, an adjacent action merely because it is long, short, at an episode edge, or contains waiting/stationary holding.
- For an explicit Idle segment, action_text should be exactly "Idle." and skill should be Idle. Mark it wrong only when clear local evidence shows that the core event of another allowed skill occurs inside it; do not invent a new boundary from approach, retreat, hovering, arm repositioning, occlusion, or waiting alone.
- Recover is valid only when a clear failed attempt, drop, loss of control, or corrective retry is the dominant action. Review existing Recover segments normally, but require visible failure/correction evidence.
- Do not add or expand Recover for an ordinary pause, hover, occlusion, transport, arm repositioning, successful action, or merely long duration. Never create multiple Recover segments from ambiguous micro-transitions.
- Pick and Place should not be collapsed into one segment. A Pick segment should not include the core release event; a Place segment should not include the core grasp/lift event.
- Only split a Pick segment into failed Pick/Place/re-Pick when local frames clearly show all of these: gripper release/opening, the manipulated instance becomes unsupported on the table or another surface, the gripper becomes empty, and a later distinct re-grasp occurs. Occlusion, hovering, dragging, near-table motion, or a short pause is not enough.
- Only change a Place destination to "table" when the held instance is visibly released on the table and remains there after gripper separation. A similar object visible on the table, or the gripper passing near the table, is not enough.
- A long segment is not a boundary error by itself. Transport, waiting, arm switching, hovering, approach, and cleanup frames are acceptable when the dominant atomic skill remains the same and no core event of another subtask occurs.
- Do not infer a boundary error by comparing one segment's duration against other segments. A Pick or Place segment may include idle, stationary holding, hovering, or settling frames after the core event when no new atomic skill starts before the next boundary.
- For the first and last segment of an episode, be especially tolerant of leading or trailing idle frames that simply cover the episode boundary. Do not require the segment to end exactly when the visible grasp/release event finishes.
- Mark an overlong tail as a boundary_error only when the tail contains the core event of another skill, or when an obvious separate atomic action begins and should be its own segment.
- Do not over-segment micro-transitions. Split only when the leading atomic skill changes.
- The first segment should begin at the episode start, the last segment should end at the episode end, and adjacent segments should be continuous.

Subtask-local description rules:
- Check object names primarily inside the current segment and its same-object Pick/Place pair. Do not compare against later unrelated objects as if they were the same item.
- For wrong_object, color, material, and wrong_bin, require clear local evidence from the current segment or same-object pair. If the evidence is ambiguous, keep the original annotation and use verdict "correct".
- When multiple similar objects exist, identify the manipulated object by gripper contact, being held, and motion continuity. Do not infer the gripper picked the wrong object from global inventory context or from a similar object still visible on the table.
- Treat duplicate or near-duplicate objects as separate physical instances. Use initial position, gripper contact point, held trajectory, and release location to track which instance is being manipulated.
- A similar white/light paper, tissue, bag, wrapper, or cloth still visible on the table after a Pick or Place does not prove that the held instance was not picked or not placed. It may be another instance.
- Do not label a Place segment as "placed back on the table" unless the held instance is visibly released onto the table. Seeing a lookalike object on the table is insufficient.
- Give higher weight to the object visibly held by the gripper during lift/transport/release than to objects still lying on the tabletop. Tabletop inventory is orientation context, not proof of what was picked.
- Before changing a color/material/object name, verify the object while it is held or released in the local Pick/Place pair. Evidence such as "the other object remains on the table" or "the table is empty" is not sufficient by itself.
- For small crumpled paper/tissue/bag/wrapper objects, avoid color swaps based on shadows, blur, or table-inventory reasoning. If the gripper-held color is not clear, keep the original annotation.
- Do not remove a Pick/Place pair because the tabletop is empty if the gripper is holding or transporting an object in that segment. A held object can be small, white, crumpled, or partially occluded.
- Failed attempts and retries are valid separate actions when the video shows a pick, drop/place back to the table, re-pick, or later release. Do not merge them away unless no held-object or release evidence exists in the local frames.
- When such a failed/corrective interval is labeled Recover, keep Recover if the local evidence supports it. Its action_text should describe the failed attempt or corrective recovery, rather than duplicate the later successful action as though both succeeded. Any newly proposed Recover must cite clear release/drop/loss-of-control and corrective-action frame evidence in review_comment.
- Do not invent failed attempts from ambiguous contact, occlusion, or hover frames. A failed attempt repair needs clear release-empty-regrasp evidence for the same instance.
- For claims such as "the gripper only hovers and does not grasp/lift," require clear before/during/after evidence that no object becomes stably held. If the gripper is holding a similar object, do not call the segment a failed attempt unless the visible evidence proves the held object is not the annotated target.
- Bin/container colors and target locations must match the current release/target evidence. Do not infer a wrong bin just because another bin is globally more visible.
- Use the official task instruction and inventory only as helpful context when provided. Inventory is not proof by itself and must not override clear local visual evidence or the annotated subtask flow.
- "Move" means robot base/chassis movement only. Arm approach/reach/hover should not be labeled as Move.

Output contract:
- Your only repair channel is corrected_action_config plus one overall review_comment.
- If the entire human action_config is correct, set verdict to "correct" and output the original action_config unchanged.
- "Unchanged" means the same segment count and the same start_frame, end_frame, action_text, and skill values. Do not paraphrase, polish, rename, or re-segment correct parts.
- If any part of the human action_config is wrong and the repair is clear, set verdict to "incorrect" and output a new full corrected_action_config that represents what you believe the correct complete annotation should be.
- The corrected_action_config must always be the full video config, not only the changed segments.
- Preserve every segment that does not need repair exactly as written; change only the segments or boundaries required by the visual evidence.
- If evidence is insufficient or the repair is not clear, set verdict to "correct" and output the original action_config unchanged.
- corrected_action_config must cover the entire video from frame 0 through frame <total_frames - 1> with continuous inclusive frame ranges.
- Every corrected segment must contain start_frame, end_frame, action_text, and skill.
- Use only allowed_skills for skill.
- review_comment should be a single overall comment explaining why corrected_action_config is identical or why it differs from the original. Mention affected segment ids and evidence frame ids when useful.
- If review_comment claims a wrong_object, failed Pick, failed Place, or returned-to-table action involving duplicate/similar objects, it must explain how the manipulated instance was tracked by contact/hold/release frames. If you cannot explain that instance trace, do not make that repair.
- Do not output issues, suggested_patch, patch_type, manual_only, or per-issue accept/reject objects.
- Set "needs_more_views": false. This run is binary: if evidence is insufficient, keep the original action_config and use verdict "correct".
- Binary mode is enabled. Do not output "uncertain". The only allowed verdict values are "correct" and "incorrect".

Return valid JSON only in this exact schema:
{
  "verdict": "correct | incorrect",
  "needs_more_views": false,
  "corrected_action_config": [
    {
      "start_frame": 0,
      "end_frame": 20,
      "action_text": "Corrected text.",
      "skill": "Pick"
    }
  ],
  "review_comment": "Brief Chinese explanation of why the corrected config is identical or different.",
  "confidence": 0.0
}
\end{lstlisting}

After the model suggestions are submitted, reviewers may only accept or reject them one by one. If both the human annotation and the model suggestion are incorrect, the case is referred to an administrator for adjudication and correction.

\begin{figure}[h]
    \centering
    \includegraphics[width=0.96\linewidth]{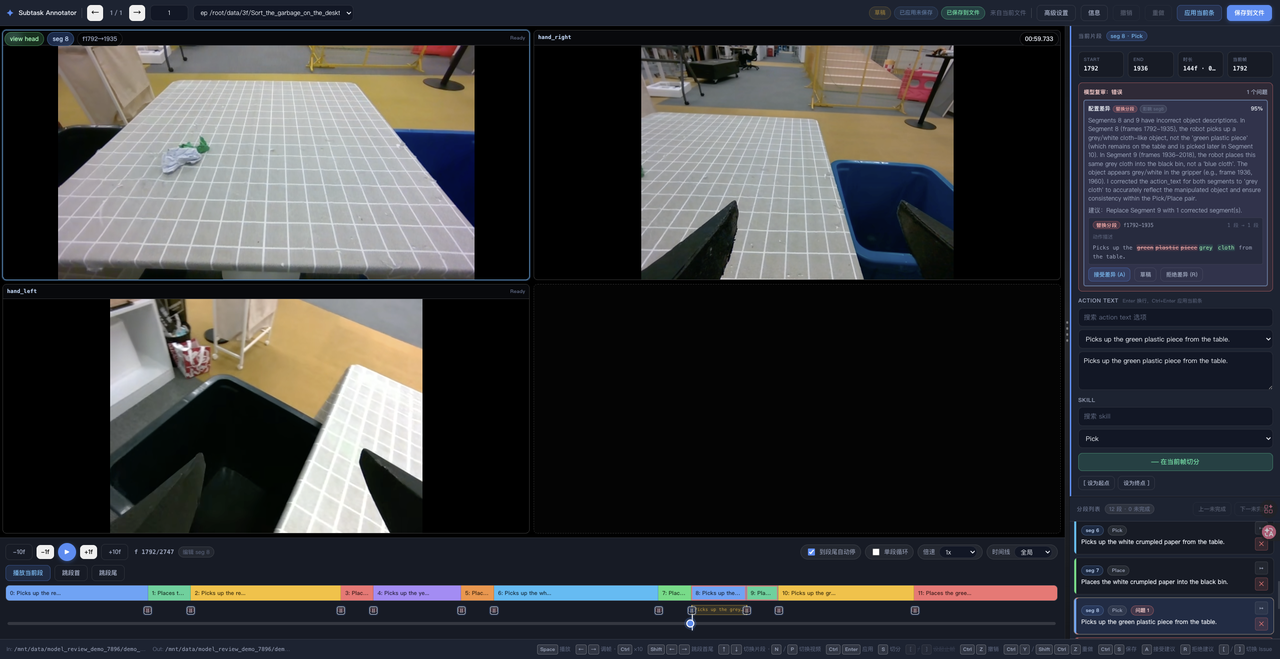}
    \caption{Model-assisted review interface.}
    \label{fig:review-interface}
\end{figure}

\subsection{RoboAtom-Data}
\label{app:roboatom-data}

The full archived version of RoboAtom-Data contains 1,446,974 trajectories and 6,876,282 segments with non-empty skill labels from 16 sources; per-source statistics are given in Table~\ref{tab:per-source}. Apart from the private data, the sources are ABC-130k~\citep{abc130k}, EBench~\citep{ebench}, Galaxea Open-World~\citep{galaxea}, InternData-A1~\citep{interndata}, LIBERO~\citep{libero}, RoboCOIN~\citep{robocoin}, RoboCasa~\citep{robocasa}, RoboCerebra~\citep{robocerebra}, RoboDojo~\citep{robodojo}, RoboMIND~\citep{robomind}, RoboMIND 2.0~\citep{robomind2}, RoboTwin 2.0~\citep{robotwin}, RH20T~\citep{rh20t}, AgiBot World Beta, and AgiBot World 2026~\citep{agibot}. The evaluation sets are RoboAtom-Bench (500 trajectories) and WGO-Bench (100 trajectories).

For Table~\ref{tab:datasets}, hours is the total trajectory (recording) duration, and $\dagger$ marks statistics we recomputed after download. Each source keeps its original annotation unit and skill taxonomy.

In Table~\ref{tab:datasets}, AgiBot World 2026 merges three real-robot themes, and Galaxea Open-World uses locally converted annotations; Table~\ref{tab:per-source} counts only the released set of this paper and its re-segmentation results. RoboInter-Data follows the official recording unit and does not deduplicate multi-view recordings; Segments for BEHAVIOR-1K 2026 is the number of logical skill annotations.

RoboAtom-Data is counted by unique trajectory identifier and by segments with non-empty skill labels. Tasks are deduplicated by ``dataset + normalized task identifier'', merging duplicate tasks across embodiments, batches, and train/validation splits; Embod. is deduplicated by the raw non-empty \texttt{robot\_type} string and denotes the number of metadata types.

\begin{table}[h]
    \centering
    \renewcommand{\thetable}{A.2}
    \caption{Per-source statistics of RoboAtom-Data. $\dagger$: local or recomputed statistics. Private is one merged source; segments require non-empty skill labels. Skills counts distinct non-empty labels per source and is non-additive. Galaxea includes retained source labels as well as core skills; the total of 29 denotes the core vocabulary size. Embod. counts distinct non-empty \texttt{robot\_type} strings, deduplicated across sources for the total.}
    \label{tab:per-source}
    \small
    \setlength{\tabcolsep}{4pt}
    \begin{tabular}{llrrrrrr}
        \toprule
        Dataset & Domain & Hours & Trajectories & Segments & Skills & Tasks & Embod. \\
        \midrule
        Private & Real & 54.7$^\dagger$ & 3,187$^\dagger$ & 20,088$^\dagger$ & 10$^\dagger$ & 7$^\dagger$ & 2$^\dagger$ \\
        ABC-130k & Real & 3,585.6$^\dagger$ & 130,740$^\dagger$ & 2,353,804$^\dagger$ & 17$^\dagger$ & 198$^\dagger$ & 1$^\dagger$ \\
        EBench & Sim & 182.9$^\dagger$ & 6,600$^\dagger$ & 42,459$^\dagger$ & 12$^\dagger$ & 26$^\dagger$ & 1$^\dagger$ \\
        Galaxea Open-World & Real & 465.9$^\dagger$ & 19,183$^\dagger$ & 124,104$^\dagger$ & 20$^\dagger$ & 169$^\dagger$ & 1$^\dagger$ \\
        InternData-A1 & Sim & 7,073.8$^\dagger$ & 621,633$^\dagger$ & 1,740,880$^\dagger$ & 13$^\dagger$ & 1,148$^\dagger$ & 1$^\dagger$ \\
        LIBERO & Sim & 3.9$^\dagger$ & 1,712$^\dagger$ & 3,910$^\dagger$ & 8$^\dagger$ & 40$^\dagger$ & 1$^\dagger$ \\
        RoboCOIN & Real & 566.5$^\dagger$ & 86,023$^\dagger$ & 387,375$^\dagger$ & 18$^\dagger$ & 352$^\dagger$ & 10$^\dagger$ \\
        RoboCasa & Sim & 178.9$^\dagger$ & 68,598$^\dagger$ & 102,737$^\dagger$ & 10$^\dagger$ & 25$^\dagger$ & 1$^\dagger$ \\
        RoboCerebra & Sim & 13.5$^\dagger$ & 985$^\dagger$ & 8,790$^\dagger$ & 10$^\dagger$ & 100$^\dagger$ & 1$^\dagger$ \\
        RoboDojo & Sim & 20.7$^\dagger$ & 3,500$^\dagger$ & 31,063$^\dagger$ & 12$^\dagger$ & 35$^\dagger$ & 1$^\dagger$ \\
        RoboMIND & Real & 191.4$^\dagger$ & 64,217$^\dagger$ & 154,824$^\dagger$ & 11$^\dagger$ & 346$^\dagger$ & 6$^\dagger$ \\
        RoboMIND 2.0 & Real & 1,128.5$^\dagger$ & 191,342$^\dagger$ & 847,591$^\dagger$ & 16$^\dagger$ & 495$^\dagger$ & 5$^\dagger$ \\
        RoboTwin 2.0 & Sim & 33.8$^\dagger$ & 27,500$^\dagger$ & 70,396$^\dagger$ & 10$^\dagger$ & 50$^\dagger$ & 1$^\dagger$ \\
        RH20T & Real & 1,132.0$^\dagger$ & 82,883$^\dagger$ & 260,586$^\dagger$ & 25$^\dagger$ & 146$^\dagger$ & 1$^\dagger$ \\
        AgiBot World Beta & Real & 2,226.6$^\dagger$ & 136,908$^\dagger$ & 715,971$^\dagger$ & 28$^\dagger$ & 181$^\dagger$ & 1$^\dagger$ \\
        AgiBot World 2026 & Real & 23.5$^\dagger$ & 1,963$^\dagger$ & 11,704$^\dagger$ & 11$^\dagger$ & 16$^\dagger$ & 1$^\dagger$ \\
        \midrule
        \textbf{RoboAtom-Data} & \textbf{Mixed} & \textbf{16,882.1$^\dagger$} & \textbf{1,446,974$^\dagger$} & \textbf{6,876,282$^\dagger$} & \textbf{29} & \textbf{3,334$^\dagger$} & \textbf{32$^\dagger$} \\
        \bottomrule
    \end{tabular}

\end{table}

\subsection{Skill Vocabulary}
\label{app:vocabulary}

RoboAtom-Data uses a 29-class core atomic-action vocabulary: Pick, Place, Handover, Move, Open, Close, Push, Pull, Press, Rotate, Screw, Plug, Pour, Fill, Scoop, Stir, Dip, Shake, Fold, Tie, Wipe, Sweep, Wash, Brush, Cut, Peel, Strike, Scan, Throw. Semantically similar actions are grouped into the same category, for example Unfold into Fold and Unscrew into Screw; the segment descriptions of such classes keep the original verb for distinction, and only the category label is normalized.

From the 29 classes we select 15, namely Close, Fold, Handover, Move, Open, Pick, Place, Plug, Pour, Press, Pull, Push, Screw, Sweep, Wipe, as the closed vocabulary for training and RoboAtom-Bench. The selection balances coverage, cross-source distinguishability, and physical recognizability: these 15 classes cover 99.2\% (6,819,538 of 6,876,282) of the skill segments in the released set, whereas the remaining labels together account for about 0.8\% of the segments and are hard to estimate stably under Skill F1@0.7, which requires exact category matching; at the same time, their semantic boundaries are clear across data sources, whereas long-tail classes (such as Stir versus Shake and Cut versus Peel) are often entangled with adjacent actions and have low cross-source annotation consistency. The long-tail classes remain in the released set for research that needs finer granularity.

For the following datasets, we use a locally deployed Qwen3.8-27B to inspect source labels and action descriptions and visually review cases requiring further assessment. Recurring issues identified through human spot checks guide improvements to source-specific mapping and correction rules. Unresolved cases undergo further review of action semantics and segmentation granularity with GPT-5.6-sol, followed by final human review and correction. Label mapping, segment splitting, and temporal interval handling differ across sources.

\paragraph{AgiBot World Beta.} We map official skill labels and accompanying action descriptions to the unified vocabulary, retaining compatible source annotations. Human feedback guides corrections to descriptions and skill mappings. Selected segmentation adjustments are supported by annotation-extension results from the same video. Where source annotations are missing or incompatible with the target granularity, we use same-task annotation extension from human-reviewed references. Existing human corrections and leading and trailing idle intervals are preserved.

\paragraph{AgiBot World 2026.} For tasks with available official segment annotations, we combine source temporal boundaries with human-reviewed, task-specific action templates. For example, the popcorn task retains three compound Pour cycles, each covering scooping and pouring, while shopping tasks follow Pick $\rightarrow$ Pick $\rightarrow$ Place $\rightarrow$ Place sequences. Other tasks use same-task annotation extension from human-reviewed references. Existing human annotations are preserved, and unannotated leading and trailing intervals are represented as idle segments. Source annotations with conflicting action order or invalid temporal boundaries are not applied.

\paragraph{Galaxea Open-World.} We normalize source labels according to their action descriptions and use visual evidence to check action categories and annotation granularity. For composite labels such as PickAndPlace, we use physical gripper features, including opening and closing transitions, to identify the boundary between picking and placing and split the segment into separate atomic actions. The results are then verified through visual and human review.

\paragraph{RH20T.} We use the skill labels, subtask descriptions, and temporal clips provided in the RoboInter version of RH20T, mapping labels with clear semantic correspondence to the unified vocabulary. For segments merged during conversion, we restore boundaries using the original temporal clips. For source segments that contain multiple atomic actions, such as multiple pick-and-place cycles involving blocks of different colors, visual review establishes the action order and segmentation granularity for correction.

\paragraph{RoboCerebra.} We map existing segment-level skill labels and action descriptions to the unified vocabulary, retain compatible annotations, and check and standardize temporal interval endpoint conventions. Segments with semantic conflicts or incompatible action granularity are reviewed and corrected against the videos, and human-confirmed corrections are incorporated.

\subsection{Training Data}
\label{app:training-data}

Training data for RoboAtom-Anno-9B are organized by the three training stages: robot-domain mid-training, supervised fine-tuning, and GRPO. This section describes their sources and sampling; training configurations and reward definitions are given in Appendix~\ref{app:training}.

\paragraph{Robot-domain mid-training.} The corpus contains 2,841,942 samples from 27 data subsets across six task categories, summarized in Table~\ref{tab:mid-training}.

\begin{table}[h]
    \centering
    \renewcommand{\thetable}{A.4}
    \caption{Task composition of robot-domain mid-training.}
    \label{tab:mid-training}
    \small
    \begin{tabular}{lrp{0.58\linewidth}}
        \toprule
        Category & Samples & Training tasks \\
        \midrule
        Locomotion & 17,185 & Navigation actions, movement direction, and target keypoint prediction \\
        Manipulation & 954 & Grasp/release actions and manipulation keypoint prediction \\
        Robot QA & 1,007,450 & Question answering on robot state, scene, and action \\
        Grounding & 1,080,004 & Localization of 2D points, target boxes, affordance regions, and manipulation trajectories \\
        Planning & 326,747 & Task decomposition, action sequence, and next-action prediction \\
        Spatial reasoning & 409,602 & Reasoning about distance, direction, relative position, and 3D relations \\
        \midrule
        \textbf{Total} & \textbf{2,841,942} & Total training samples of the six categories \\
        \bottomrule
    \end{tabular}
\end{table}

The mid-training data sources and task descriptions are as follows. Locomotion comes from VeBrain~\citep{vebrain}; from the robot's egocentric image, it judges the movement direction, the next action, and the target landing point, and outputs navigation actions and 2D keypoints.

Manipulation: also from VeBrain. From the image, it judges the next grasp or release action and predicts manipulation keypoints. It is not full-trajectory segmentation data.

Robot QA: includes OWMM-VLM~\citep{owmm}, Robo2VLM~\citep{robo2vlm}, and RoboTwinQA. The tasks cover gripper-state question answering, scene understanding, target search, the next robot action, and spatial and grounding question answering.

Grounding: includes RefSpatial-3D and RefSpatial-Simulator~\citep{roborefer}, affordance, bounding box, PACO-LVIS~\citep{paco}, PixMo points~\citep{molmo}, and trajectory, among others. The outputs are mainly 2D points, target boxes, affordance regions, or manipulation trajectories on the image plane; this is spatial grounding and contains no subtask temporal boundaries.

Planning: includes EgoCOT~\citep{embodiedgpt}, EgoPlan~\citep{egoplan}, MuEP~\citep{muep}, and EmbodiedBench-ALFRED/Habitat~\citep{embodiedbench}, among others. From an image or a frame sequence, it generates a task plan, an action sequence, or the next action, mainly training step decomposition and action ordering.

Spatial reasoning: includes SR-91K~\citep{spacer}, VSI-100K~\citep{vsi100k}, and VLM-3R~\citep{vlm3r}, and trains spatial reasoning about distance, direction, relative position, and 3D scene relations.

\paragraph{Supervised fine-tuning.} The SFT data of the RoboAtom annotation model come from the training pool organized during supervision construction. We sample part of the tasks to build the training subset and exclude tasks that overlap with RoboAtom-Bench. The final model performs SFT with this list after robot-domain mid-training. The evaluation data for PhysExtend are described separately in Section~\ref{sec:extension} and Appendix~\ref{app:rule-extension}. Training settings are given in Appendix~\ref{app:sft}.

The open-source data control comes from AgiBot World Beta, AgiBot World 2026, and Galaxea Open-World. The three sets are first sampled by task separately and then merged; because their skill and subtask definitions are inconsistent with the evaluation closed set, semantically determinable aliases are then mapped to the closed set, samples that cannot be uniquely mapped or are ambiguous are deleted, and the result is reviewed by Qwen3-VL-235B-A22B-Instruct. The three sources total 425 tasks with at most 20 trajectories per task; after alignment, 261 tasks and 4,459 trajectories remain. The resulting Skill F1@0.7 is 17.1 (Section~\ref{sec:anno}).

\paragraph{GRPO.} Reward optimization uses a separately organized list of 5,000 trajectories. Difficulty is divided according to the offline prediction quality on the training supervision pool, and samples are stratified by source, view, and duration, keeping 3,000 medium, 1,250 hard, and 750 easy trajectories. Reward and training settings are given in Appendix~\ref{app:grpo}.

\subsection{Test-Set Distribution}
\label{app:test-distribution}

Section~\ref{sec:benchmark} defines RoboAtom-Bench. This appendix reports its full distribution: 500 trajectories and 4,018 ground-truth segments, with a median of 5 segments. Short, Medium, and Long are divided by the number of ground-truth segments, with ranges of 1--4, 5--8, and 9--62, respectively; Multiview separately evaluates annotation ability with multi-view input. The trajectory counts and detailed statistics of each subset are given in Table~\ref{tab:test-distribution}.

\begin{table}[h]
    \centering
    \renewcommand{\thetable}{A.5}
    \caption{Test-set distribution of RoboAtom-Bench. Multi-view durations are not multiplied by the number of cameras; trajectory and video counts are reported separately.}
    \label{tab:test-distribution}
    \small
    \setlength{\tabcolsep}{3pt}
    \resizebox{\linewidth}{!}{%
    \begin{tabular}{lrrrrrrrrr}
        \toprule
        Test subset & Trajectories & Share & GT segments & Mean GT segments & Median segments & Segment range & Total frames & Duration (hours) & Videos \\
        \midrule
        Short & 200 & 40.0\% & 411 & 2.06 & 2 & 1--4 & 74,214 & 0.732 & 200 \\
        Medium & 100 & 20.0\% & 634 & 6.34 & 6 & 5--8 & 126,762 & 1.193 & 100 \\
        Long & 100 & 20.0\% & 2,302 & 23.02 & 21 & 9--62 & 298,015 & 2.761 & 100 \\
        Multiview & 100 & 20.0\% & 671 & 6.71 & 5 & 1--48 & 108,105 & 1.119 & 321 \\
        \midrule
        \textbf{Total} & \textbf{500} & \textbf{100.0\%} & \textbf{4,018} & 8.04 & 5 & 1--62 & \textbf{607,096} & \textbf{5.806} & \textbf{721} \\
        \bottomrule
    \end{tabular}%
    }

\end{table}

\subsection{Language-Space Coverage}
\label{app:language-space}

\begin{figure}[p]
    \centering
    \includegraphics[width=0.96\linewidth]{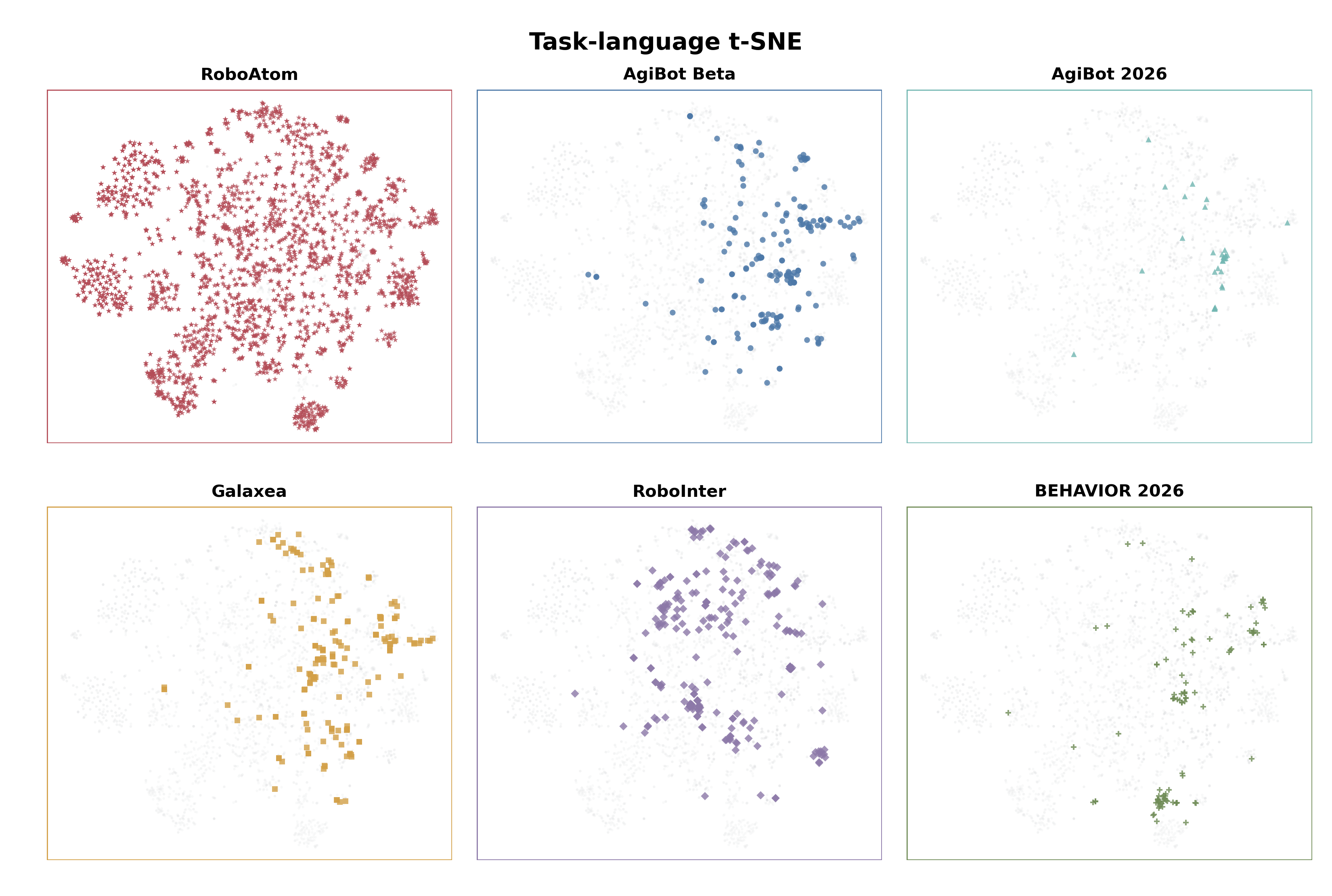}
    \caption{Task-instruction distributions across datasets. Colored points highlight each dataset within the shared t-SNE projection.}
    \label{fig:task-language}
\end{figure}

\begin{figure}[p]
    \centering
    \includegraphics[width=0.96\linewidth]{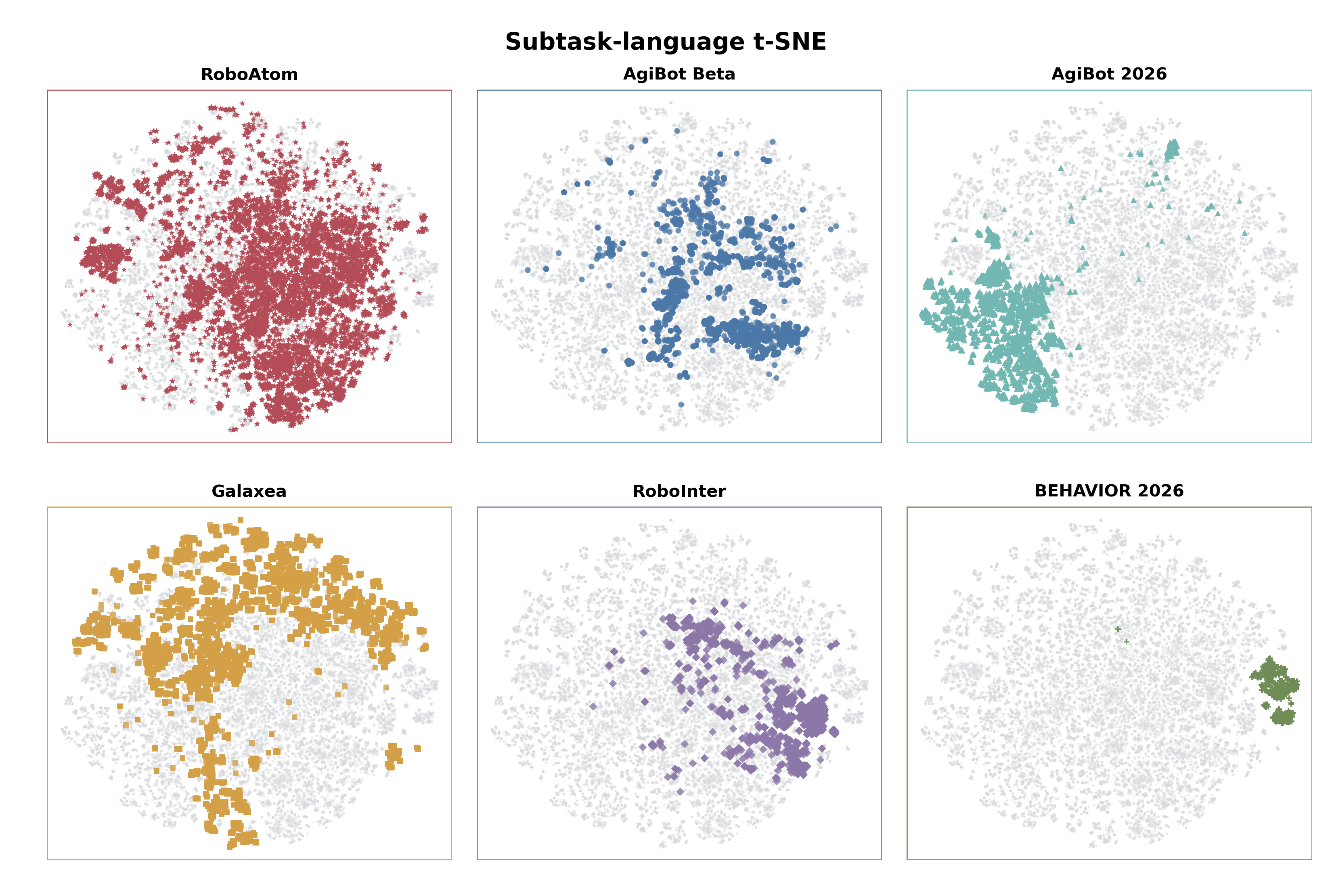}
    \caption{Action-description distributions across datasets, illustrating the broad atomic-action coverage of RoboAtom-Data.}
    \label{fig:subtask-language}
\end{figure}

RoboAtom-Data covers a broad range of task instructions and action descriptions across manipulation scenarios. Figures~\ref{fig:task-language} and~\ref{fig:subtask-language} provide dataset-specific views of the distributions summarized in Figure~\ref{fig:data-characteristics}(e--f).

\clearpage

\section{Training and Evaluation Settings}
\label{app:training}

RoboAtom-Anno-9B follows the three-stage method in Section~\ref{sec:anno-method}: robot-domain mid-training (Appendix~\ref{app:mid-training}), supervised fine-tuning on a task-sampled subset (Appendix~\ref{app:sft}), and GRPO (Appendix~\ref{app:grpo}). The downstream policy in Section~\ref{sec:policy-method} fine-tunes $\pi_{0.5}$ with OpenPI (Appendix~\ref{app:executor}). Benchmark construction is described in Section~\ref{sec:benchmark-construction}, and evaluation settings are given in Section~\ref{sec:anno} and Appendix~\ref{app:eval-metrics}.

\subsection{Supervised Fine-Tuning Settings}
\label{app:sft}

Supervised fine-tuning is performed in LLaMA-Factory~\citep{llamafactory} with Qwen3.5-9B~\citep{qwen35} as the base; the construction and task split of the training subset are given in Appendix~\ref{app:training-data}. The visual update-scope ablation is not preceded by mid-training; the final model is initialized from the robot-domain mid-training weights and then goes through supervised fine-tuning and GRPO optimization on this subset. The LoRA~\citep{lora} rank is 8, applied to the linear layers on both the language and vision sides, and the visual merger is updated as well. The language-side LoRA is trained in all four visual update ablations; the switches for ViT-LoRA and the visual merger module are given in Table~\ref{tab:visual-ablation}.

Videos are sampled at 2 FPS with at most 448 frames per trajectory, using the Qwen3.5 chat template without chain of thought, and segment boundaries are output by frame index. The effective batch size is 64 (8 GPUs $\times$ 2 per GPU $\times$ 4 gradient accumulation steps), the learning rate is $2\times10^{-5}$, and training runs for 500 optimizer steps.

\subsection{Robot-Domain Mid-Training}
\label{app:mid-training}

Before supervised fine-tuning, RoboAtom-Anno-9B undergoes robot-domain mid-training in LLaMA-Factory. The data sources and task composition are described in Appendix~\ref{app:training-data} and Table~\ref{tab:mid-training}.

Training is initialized from Qwen3.5-9B and runs for one epoch with a learning rate of $2\times10^{-5}$, a per-GPU batch size of 8, and gradient accumulation of 4; videos are still input at 2 FPS with at most 448 frames; the LoRA rank is 8, and the visual merger module is updated.

\subsection{GRPO Reward and Training Settings}
\label{app:grpo}

GRPO~\citep{grpo} is implemented in ms-swift~\citep{swift} and continues optimization from the model that has completed robot-domain mid-training and supervised fine-tuning. The reward-training data and difficulty-based sampling are described in Appendix~\ref{app:training-data}.

\begin{equation}
    R=0.70R_{\mathrm{temporal}}+0.10R_{\mathrm{soft}}+0.18R_{\mathrm{edit}}+0.02R_{\mathrm{format}}
    \label{eq:grpo-reward}
\end{equation}

$R_{\mathrm{temporal}}$ is the F1@0.5 based on ordered temporal matching and does not require correct skill labels; $R_{\mathrm{soft}}$ is the sum of IoU over ordered matches divided by the larger of the numbers of predicted and GT segments, which provides continuous feedback for boundary changes; $R_{\mathrm{edit}}$ is the edit similarity of the skill sequence; and $R_{\mathrm{format}}$ checks strict parsing and field constraints. If the output cannot be parsed, the first three terms are set to zero, the format term is set to $-1$, and the total reward is $-0.02$.

GRPO uses 16 GPUs with a training batch size of 1 per GPU and gradient accumulation of 4; 8 candidate outputs are sampled per prompt, each prompt is updated 2 times, and a group of candidates is generated every 4 steps. The learning rate is $5\times10^{-6}$ with cosine decay; the KL coefficient is 0.02, the clipping range is 0.2, and the maximum generation length is 4,096 tokens. Videos are sampled at 2 FPS with at most 448 input frames. The candidate advantage is computed as the group-centered reward divided by the within-group standard deviation:

\begin{equation}
    A_i=\frac{r_i-\bar r}{s_r+10^{-4}}
    \label{eq:grpo-advantage}
\end{equation}

% Compute provenance: mid-training run qwen35-9b-pretrain-0901r8f48,
% formal/output/all_results.json, train_runtime = 231307.0138 seconds (48 GPUs);
% final SFT model Qwen3.5-9B-cortex-frame-n-singleview-s500-20260904/all_results.json,
% train_runtime = 12477.7044 seconds (8 GPUs).
% GRPO model Qwen3.5-9B-cortex-frame-n-singleview-s500-grpo-original5000-ep1-16gpu-aligned-20260906,
% v0-20260907-022419/logging.jsonl, final train_runtime = 82010.1297 seconds (16 GPUs).
% The reported GRPO compute covers the full run, including rollout generation.

\subsection{Downstream Executor}
\label{app:executor}

The downstream policies for LIBERO, LIBERO-Plus, and RoboTwin all fully fine-tune $\pi_{0.5}$ with OpenPI~\citep{openpi} using a shared global batch size of 256. LIBERO uses 8 GPUs and 30,000 training steps; RoboTwin uses 32 GPUs (4 nodes $\times$ 8 GPUs) and 60,000 training steps. The language condition of the task-level baseline is the complete task instruction for both training and inference, without the high-level scheduling of System~2 or the task-plan prior; the training language condition of our method is the skill instruction corresponding to each atomic-action segment. At inference, our method receives the atomic-action instructions issued by System~2 according to observations and the execution history.

System~2 uses Qwen3.8-27B as the planner. A task-plan prior from the task knowledge base provides candidate skills and action descriptions; the planner combines this prior with visual observations and execution history to retain the current atomic-action instruction or select another. System~1 executes the selected instruction through robot action chunks.

The comparison of the $\pi_{0.5}$ baseline with augmentation on and off is given in Appendices~\ref{app:libero-aug}--\ref{app:robotwin-aug}.

Each RoboTwin evaluation has 5,000 trials: the dual system succeeds 4,101 times under Clean and 2,516 times under Random, 6,617/10,000 in total, for an average of 66.17\%; the end-to-end policy succeeds 4,035 times under Clean and 2,402 times under Random, 6,437/10,000 in total, for an average of 64.37\%.

\subsection{Visual Input and Evaluation Metrics}
\label{app:eval-metrics}

The main results use the 500 fixed human-annotated test trajectories of Appendix~\ref{app:test-distribution}. RoboAtom-Bench uses the GT-active evaluation window, the time interval covered by the standard annotation, with no background skill labels excluded. Section~\ref{sec:benchmark-metrics} defines mIoU, temporal F1@0.7, Skill F1@0.7, and Edit Score. Table~\ref{tab:benchmark-reference} reports quality and cost for the reference models and RoboAtom-Anno-9B.

For mIoU, an order-preserving one-to-one alignment maximizes total temporal IoU, independently of skill labels. The aligned IoUs are summed and divided by the number of GT segments; unmatched GT contributes zero. Temporal F1 uses a separate ordered alignment that first maximizes the number of pairs meeting the IoU threshold and then their total IoU. Skill F1 follows MS-TCN-style segmental matching~\citep{mstcn}: predictions are processed in temporal order, and each is compared with all GT segments of the same skill without an order constraint. A prediction is a true positive if its highest-IoU same-skill GT segment meets the threshold and has not already been matched; otherwise it is a false positive, without falling back to the next-best GT segment. Each F1 metric therefore constructs its own set of matched pairs. The formal temporal thresholds are 0.3, 0.5, and 0.7; the main tables report 0.7. Both F1 metrics accumulate TP, FP, and FN over episodes before computing Micro F1.

Edit uses normalized exact skill sequences with adjacent identical labels collapsed. Levenshtein distance assigns unit cost to insertion, deletion, and substitution; normalization is by the longer sequence. We report the arithmetic mean of episode scores; all other metrics pool segments over episodes. All metrics are reported on a 0--100 scale. An empty prediction receives zero Edit when the GT sequence is nonempty, and empty predictions remain failures in the temporal and skill metrics. Edit ignores timestamps and may also ignore splits or merges between consecutive instances of the same skill.

The base Qwen3.5-9B follows the input format in Section~\ref{sec:seed}. Frames are uniformly sampled across each video at a target rate of 2 FPS, including both endpoints and capped at 448 frames. Each sampled frame carries its source frame index. Single-view trajectories use one camera; synchronized multiview trajectories use a mosaic at each sampled time. Multiview is a disjoint trajectory subset with its own task and source composition. Reference WGO evaluation uses the full video; RoboAtom-Bench uses the GT-active interval. Empty predictions are retained in all reported reference-model scores.

For visual input, RoboAtom-Anno-9B and its SFT/GRPO configurations are evaluated at 2 FPS with at most 448 frames and deterministic decoding.

WGO-Bench has 100 trajectories, of which 75 are Robot (50 DROID, 25 Galaxea) and 25 are Ego (HomER). We use a unified evaluator that covers the whole video without removing background intervals and report mIoU and F1@0.7 for temporal overlap and segment matching. WGO uses open-vocabulary annotation and has no standard labels corresponding to our unified categories, so Skill F1@0.7 is not reported.

The evaluation closed set is Close, Fold, Handover, Move, Open, Pick, Place, Plug, Pour, Press, Pull, Push, Screw, Sweep, Wipe. RoboAtom-Anno-9B outputs a JSON object in which \texttt{segments} is an ordered list of segments, each containing \texttt{skill}, \texttt{subtask}, \texttt{start\_frame}, and \texttt{end\_frame}; start and end frames use zero-based closed-interval coordinates of the original video.

\section{Supplementary Experimental Results}
\label{app:results}

\subsection{Update Scope of the Visual Modules}
\label{app:visual-update}

We examine the scope of SFT parameter updates on the task-sampled RoboAtom training subset. Updating both ViT-LoRA and the visual merger gives the highest Skill F1@0.7 in this ablation, 22.7 (Table~\ref{tab:visual-ablation}). We select this configuration for subsequent training. For the main-text results, training annotations undergo a second review to ensure consistent within-task action sequences.

\begin{table}[h]
    \centering
    \renewcommand{\thetable}{C.1}
    \caption{Ablation of the update scope of the visual modules. All configurations update the language-side LoRA for 500 steps on the same task-sampled RoboAtom subset.}
    \label{tab:visual-ablation}
    \small
    \begin{tabular}{lccccc}
        \toprule
        Configuration & ViT-LoRA & visual.merger & mIoU & F1@0.7 & Skill F1@0.7 \\
        \midrule
        \ding{172} Language-only & Off & Frozen & 44.3 & 28.7 & 21.7 \\
        \ding{173} + merger & Off & Updated & \textbf{50.0} & 28.7 & 21.4 \\
        \ding{174} + ViT-LoRA & On & Frozen & 43.8 & 28.8 & 22.3 \\
        \ding{175} + ViT-LoRA + merger & On & Updated & 48.7 & \textbf{29.5} & \textbf{22.7} \\
        \bottomrule
    \end{tabular}

\end{table}

\subsection{Analysis of Annotation-Model Training Configurations}
\label{app:training-config}

Robot-domain mid-training uses 2,841,942 samples from 27 subsets, covering six task categories: robot question answering, spatial grounding, spatial reasoning, planning, locomotion, and manipulation. After one epoch of domain training from Qwen3.5-9B, with the same task-sampled supervision and 500-step SFT configuration, Skill F1@0.7 rises from 24.2 to 26.9 and mIoU from 52.9 to 54.6. Continuing with RLVR (GRPO) raises Skill F1@0.7 to 30.1 and temporal F1@0.7 from 35.5 to 41.0, and brings the predicted-to-GT segment-count ratio from 1.20 to 0.99. The mid-training data composition is given in Appendix~\ref{app:training-data}, and the training hyperparameters are given in Appendices~\ref{app:mid-training}, \ref{app:sft}, and~\ref{app:grpo}.

\subsection{Annotation Difficulty of the Test Subsets}
\label{app:subset-difficulty}

The detailed subset breakdown is given in Table~\ref{tab:subset-difficulty}; Skill F1@0.7 decreases as the number of segments grows, from 71.1 on the Short subset to 17.3 on the Long subset.

\begin{table}[h]
    \centering
    \renewcommand{\thetable}{C.3}
    \caption{Skill F1@0.7 of RoboAtom-Anno-9B on each RoboAtom-Bench test subset.}
    \label{tab:subset-difficulty}
    \small
    \begin{tabular}{lccc}
        \toprule
        Subset & Trajectories & Mean GT segments & Skill F1@0.7 \\
        \midrule
        Short & 200 & 2.06 & 71.1 \\
        Medium & 100 & 6.34 & 43.7 \\
        Long & 100 & 23.02 & 17.3 \\
        Multiview & 100 & 6.71 & 33.9 \\
        Overall & 500 & 8.04 & 30.1 \\
        \bottomrule
    \end{tabular}
\end{table}

\subsection{PhysExtend Implementation and Feature Evidence}
\label{app:rule-extension}

\begin{figure}[h]
    \centering
    \includegraphics[width=0.96\linewidth]{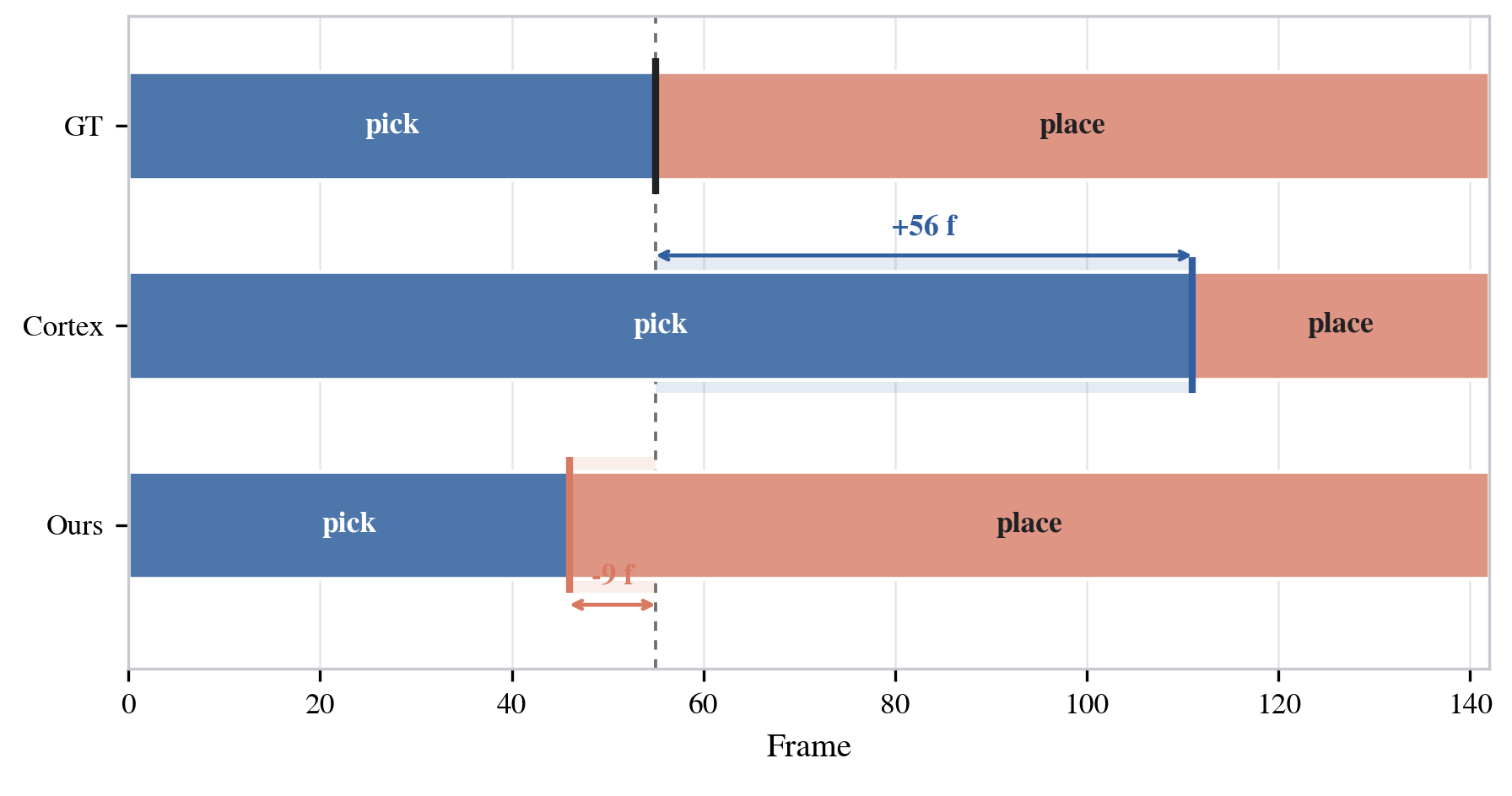}
    \caption{Boundary localization of Cortex and PhysExtend compared with ground truth (GT).}
    \label{fig:boundary-comparison}
\end{figure}

Boundary localization jointly considers relative position, cumulative motion progress, local feature differences, segment duration, and skill-dependent gripper events; the objective is given in Section~\ref{sec:extension}. Candidate boundaries are generated from reference durations, cumulative motion alignment, and gripper events, and their positions are determined by joint dynamic programming over ordered candidates before correction with reliable Pick/Place events. Each reference trajectory generates boundaries independently; corresponding boundaries are fused and then monotonically projected so that the segments do not overlap and cover the whole trajectory, preserving the shared action structure while reducing sensitivity to noise in any single reference trajectory. We characterize Pick and Place by gripper-state transitions (Figure~\ref{fig:physical-evidence}(a)), with event direction calibrated from the reference trajectories to accommodate different robot encodings. Figure~\ref{fig:physical-evidence}(b) illustrates the chassis-motion evidence. Bimanual coordination, contact, pauses, rotation, and direction changes provide additional cues for skill boundaries, and Figure~\ref{fig:physical-evidence}(c) summarizes the qualitative associations between the 15 primary skills and these physical features. Boundary and within-segment evidence is compared in Appendix Figure~\ref{fig:feature-response}. The extension evaluation contains 100 tasks with five annotated trajectories per task; we evaluate reference counts from one to four and aggregate results across all reference combinations. Figure~\ref{fig:boundary-comparison} compares the boundary timelines of Cortex and PhysExtend with ground truth.

\begin{figure}[h]
    \centering
    \includegraphics[width=0.96\linewidth]{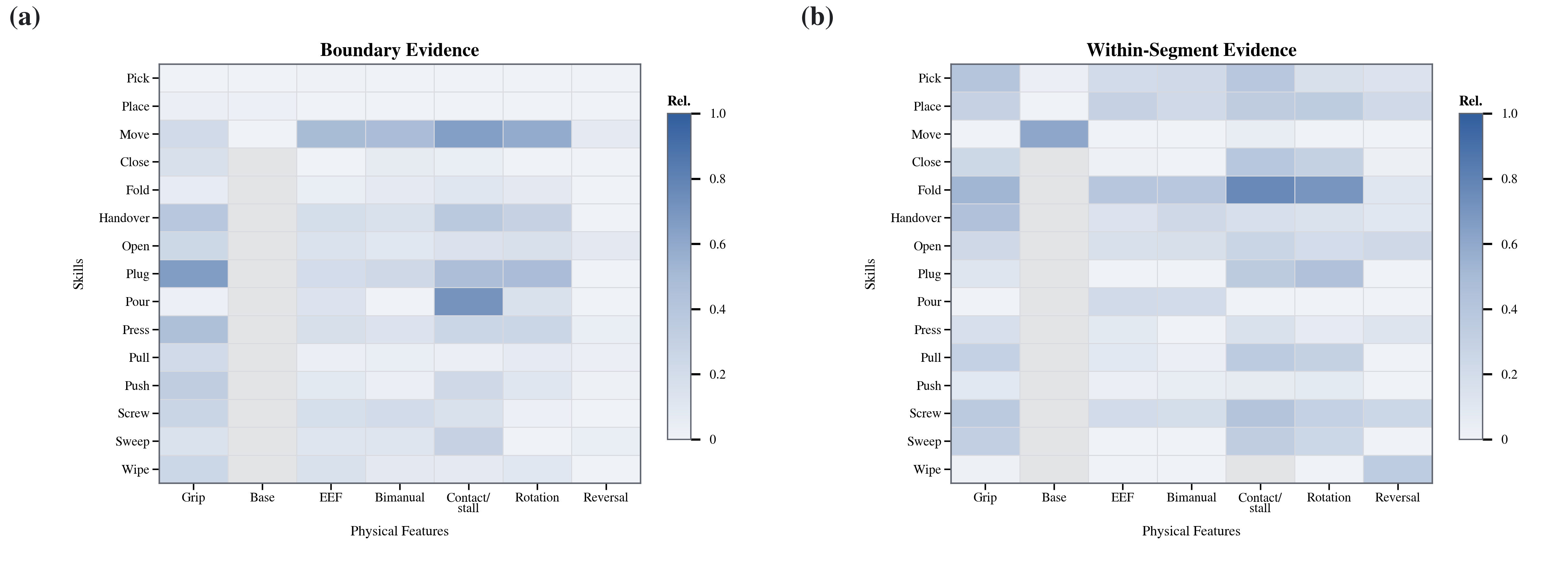}
    \caption{Response of physical features at action boundaries versus within segments. (a) Relative evidence at the boundaries; (b) evidence of the same features within the segments (at the center).}
    \label{fig:feature-response}
\end{figure}

With a single CPU and 1 reference trajectory, PhysExtend processes 15,952 trajectories per hour, about 17.1 times the throughput of Cortex. The corresponding compute costs for Cortex and PhysExtend are 0.005253 and 0.000307 USD per 1,000 trajectories, respectively, or approximately 0.000481 and 0.000028 USD per source-video hour, respectively.

\subsection{Image Augmentation Control for $\pi_{0.5}$}
\label{app:libero-aug}

With image augmentation turned off, the LIBERO-Long success rate of $\pi_{0.5}$ is 95.00\%, 2.60 percentage points higher than the 92.40\% with augmentation on; the standard LIBERO Overall results are 97.125\% and 96.85\%, respectively. The OpenPI maintainers also explain in \href{https://github.com/Physical-Intelligence/openpi/issues/442#issuecomment-2817401860}{issue \#442} that the augmentation defaults of $\pi_0$ and $\pi_0$-FAST at the time came from historical settings and were kept during fine-tuning for consistency with the respective pretraining. Given this background and the measured performance on long-horizon tasks, the main text uniformly compares $\pi_{0.5}$ and our method with image augmentation turned off.

Tables~\ref{tab:libero-standard} and~\ref{tab:libero-plus-aug} compare $\pi_{0.5}$ with image augmentation off (w/o aug) and on (w/ aug).

\begin{table}[h]
    \centering
    \renewcommand{\thetable}{C.5(a)}
    \caption{Image augmentation control of the $\pi_{0.5}$ baseline on standard LIBERO (\%).}
    \label{tab:libero-standard}
    \small
    \begin{tabular}{lccccc}
        \toprule
        Method & Spatial & Object & Goal & Long & Overall \\
        \midrule
        $\pi_{0.5}$ (w/o aug) & 97.00 & 99.00 & 97.50 & 95.00 & 97.125 \\
        $\pi_{0.5}$ (w/ aug) & 98.80 & 98.20 & 98.00 & 92.40 & 96.85 \\
        \bottomrule
    \end{tabular}
\end{table}

\begin{table}[h]
    \centering
    \renewcommand{\thetable}{C.5(b)}
    \caption{Image augmentation control of the $\pi_{0.5}$ baseline on LIBERO-Plus (\%).}
    \label{tab:libero-plus-aug}
    \small
    \begin{tabular}{lccccc}
        \toprule
        Method & Spatial & Object & Goal & Long & Overall \\
        \midrule
        $\pi_{0.5}$ (w/o aug) & 93.01 & 85.23 & 79.93 & 78.56 & 84.05 \\
        $\pi_{0.5}$ (w/ aug) & 91.63 & 89.36 & 79.08 & 79.48 & 84.77 \\
        \bottomrule
    \end{tabular}
\end{table}

\subsection{Image Augmentation Control on RoboTwin}
\label{app:robotwin-aug}

Table~\ref{tab:robotwin-aug} compares the success rates of the $\pi_{0.5}$ baseline on RoboTwin with image augmentation on and off. w/ aug and w/o aug denote image augmentation on and off. Clean and Random denote evaluation in the clean and randomized settings after training on clean data, and Avg. is their arithmetic mean.

\begin{table}[h]
    \centering
    \renewcommand{\thetable}{C.6}
    \caption{Image augmentation control of the $\pi_{0.5}$ baseline on RoboTwin (\%).}
    \label{tab:robotwin-aug}
    \small
    \begin{tabular}{lccc}
        \toprule
        Method & Clean & Random & Avg. \\
        \midrule
        $\pi_{0.5}$ (w/ aug) & 70.70 & 46.00 & 58.35 \\
        $\pi_{0.5}$ (w/o aug) & 80.70 & 48.04 & 64.37 \\
        \bottomrule
    \end{tabular}
\end{table}

\section{Inference Efficiency and Cost}
\label{app:cost}

This appendix describes the measurement and cost accounting used for Table~\ref{tab:benchmark-reference}. Costs cover inference only and exclude model training and data construction.

\subsection{Measurement and Cost Accounting}
\label{app:gpu-cost}

Inference cost is reported in USD per source-video hour. For API models, we calculate costs from actual input, cached-read, and output token usage at official API prices, including format repairs and failed requests, and divide by the total source-video duration. For local models, we calculate costs from the GPU hours actually occupied during the measurement and a common RTX 4090 rental rate. An eight-GPU machine rented for 2,235 USD per month, assuming 720 hours per month, gives a unit price of 0.3880208 USD per GPU hour. Thus, the cost per source-video hour is the unit GPU-hour price multiplied by the occupied GPU hours and divided by the processed source-video hours.

Timing spans video processing through annotation generation, including video decoding, frame extraction, multi-view camera composition where applicable, visual encoding, and generation; model loading and warm-up are excluded. Concurrent workers are measured over the shared wall-clock interval from the earliest start to the latest finish, accounting for the GPUs occupied during that interval. For repeated measurements, processed video duration and occupied GPU hours are accumulated consistently across passes. Synchronized multi-view inputs count as one trajectory per pass, without multiplying duration by camera count. Storage, data transfer, and labor costs are excluded. Costs are calculated from unrounded measurements and reported to three decimal places in Table~\ref{tab:benchmark-reference}.

\end{document}